\documentclass[10pt,twocolumn,letterpaper]{article}

\usepackage[accsupp]{axessibility} 
\usepackage[pagenumbers]{iccv} 

\definecolor{iccvblue}{rgb}{0.21,0.49,0.74}
\usepackage[pagebackref,breaklinks,colorlinks,allcolors=iccvblue]{hyperref}
\usepackage{colortbl}
\usepackage{graphicx} 
\usepackage{caption}
\usepackage{multirow}
\usepackage{soul}
\usepackage{tikz}
\usepackage{adjustbox}
\newcommand{\closeruline}[1]{%
  \begin{tikzpicture}[baseline=(textnode.base)]
    \node[inner sep=0.5pt,outer sep=0.5pt] (textnode) {#1};
    \draw[yshift=0ex] (textnode.south west) -- (textnode.south east);
  \end{tikzpicture}%
}

\usepackage{tabularx}

\definecolor{darkgreen}{rgb}{0.0, 0.5, 0.0}

\usepackage{amsmath}

\title{DisTime: Distribution-based Time Representation for \\ Video Large Language Models }

\author{
\textbf{Yingsen Zeng\textsuperscript{1}\footnotemark[1]~}, 
\quad
\textbf{Zepeng Huang\textsuperscript{1}\footnotemark[1]~}, 
\quad
\textbf{Yujie Zhong\textsuperscript{1}\footnotemark[2]~},  
\quad
\textbf{Chengjian Feng\textsuperscript{1}}, 
\quad
\textbf{Jie Hu\textsuperscript{1}},  \\
\textbf{Lin Ma\textsuperscript{1}}, 
\quad
\textbf{Yang Liu\textsuperscript{2}} \\
\textbf{$^{1}$} Meituan Inc.\\
\textbf{$^{2}$} Wangxuan Institute of Computer Technology, Peking University \\
\tt\footnotesize \{yingsen\_2, huang\_zpeng\}@163.com
\; 
\tt\footnotesize jaszhong@hotmail.com 
\;
\tt\footnotesize yangliu@pku.edu.cn\\
}

\begin{document}
\maketitle

\let\thefootnote\relax\footnotetext{
\noindent $^*$ Equal contribution with random order. $^\dagger$ Corresponding author.}

\begin{abstract}

Despite advances in general video understanding, Video Large Language Models (Video-LLMs) face challenges in precise temporal localization due to discrete time representations and limited temporally aware datasets. Existing methods for temporal expression either conflate time with text-based numerical values, add a series of dedicated temporal tokens, or regress time using specialized temporal grounding heads. 
To address these issues, we introduce \textbf{DisTime}, a lightweight framework designed to enhance temporal comprehension in Video-LLMs.
DisTime employs a learnable token to create a continuous temporal embedding space and incorporates a \textbf{Distribution-based Time Decoder} that generates temporal probability distributions, effectively mitigating boundary ambiguities and maintaining temporal continuity.
Additionally, the \textbf{Distribution-based Time Encoder} re-encodes timestamps to provide time markers for Video-LLMs.
To overcome temporal granularity limitations in existing datasets, we propose an automated annotation paradigm that combines the captioning capabilities of Video-LLMs with the localization expertise of dedicated temporal models.
This leads to the creation of \textbf{InternVid-TG}, a substantial dataset with 1.25M temporally grounded events across 179k videos, surpassing ActivityNet-Caption by 55 times. Extensive experiments demonstrate that DisTime achieves state-of-the-art performance across benchmarks in three time-sensitive tasks while maintaining competitive performance in Video QA tasks. DisTime is released at \url{https://github.com/josephzpng/DisTime}.
\vspace{-8mm}
\end{abstract}    
\section{Introduction}
\label{sec:intro}

With Large Language Models (LLMs) displaying powerful comprehension abilities, significant efforts have been made to bridge the gap between language and vision, enabling LLMs to tackle visual tasks. As a result, Image-LLMs~\cite{chen2024far,chiang2023vicuna,li2023blip,liu2023visual,zhu2023minigpt,pi2024perceptiongpt} have achieved breakthroughs in image-related tasks. In the realm of video tasks, which contain richer information, numerous Video-LLMs~\cite{chen2024far,li2023videochat,zhang2023video,xu2024pllava,yao2024minicpm,li2024llava} have also made substantial progress in areas such as reasoning, captioning, and summarization. However, for temporal comprehension, these models typically manage temporal information by capturing the sequence of events rather than the precise timing of their occurrence. This leads to significant shortcomings in time-sensitive tasks, such as moment retrieval~\cite{ gao2017tall, krishna2017dense,grauman2022ego4d,shi2022react,shi2023temporal,shi2023tridet,li2024detal}, dense video captioning~\cite{krishna2017dense,zhou2018towards}, and grounded video question answering~\cite{xiao2024can,chen2024cg}, \etc.

\begin{figure}[t]
  \centering
  \includegraphics[width=.95\linewidth]{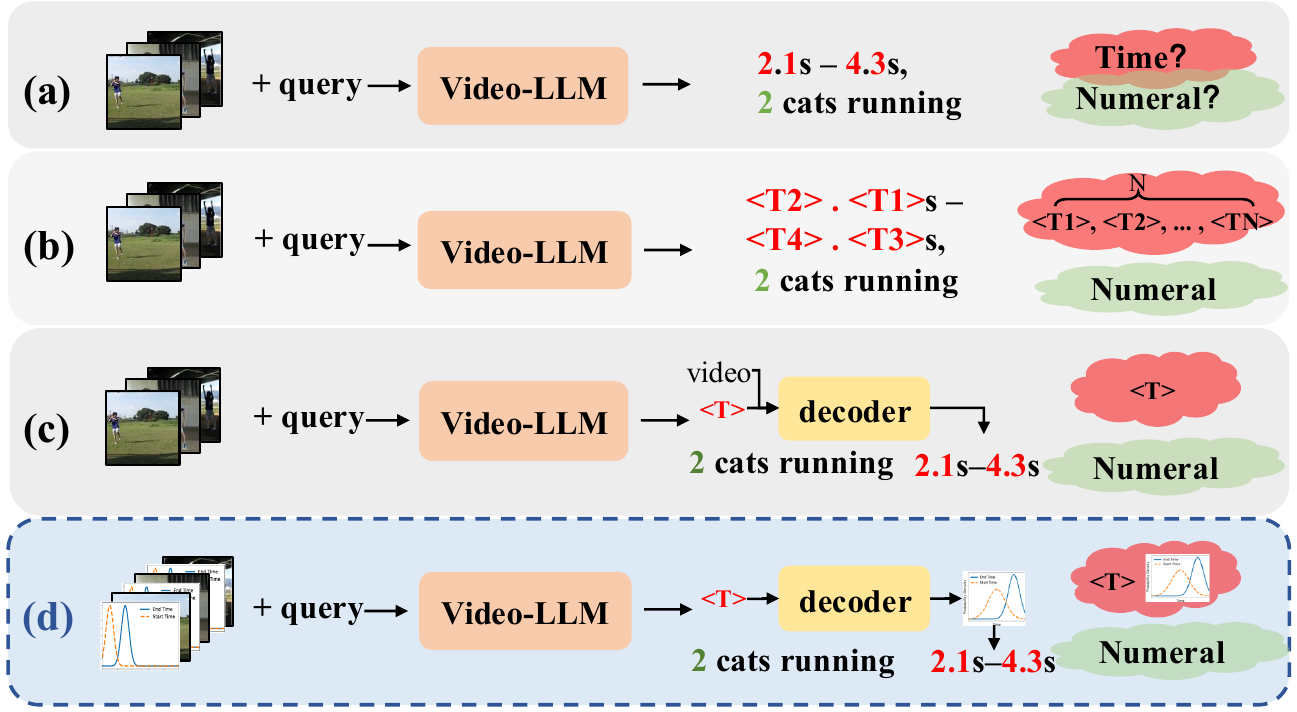}
  \vspace{-2mm}
  \caption{
  Comparison of temporal expression paradigms in Video-LLMs: (a) Text-Modal Discretization: time represented via text/number hybrids, (b) Multi-Token Discretization: dedicated temporal tokens, (c) Specialized Temporal Heads: utilizing temporal grounding models for direct regression, and (d) \textbf{DisTime}: distribution-based time decoder for temporal localization.
  }
  \vspace{-7mm}
  \label{fig:intro_sub1}
\end{figure}

\emph{Temporal expression} poses a significant challenge in time-sensitive Video-LLMs. The simplest approach is to express time using numbers, either alone or with markers, within the text modality. Models like GroundingGPT~\cite{li2024groundinggpt}, VTimeLLM~\cite{huang2024vtimellm}, and TimeMarker~\cite{chen2024timemarker} employ this strategy. However, these methods force time and numbers to share the same decision boundary, complicating the classification process, as illustrated in Fig.~\ref{fig:intro_sub1}a.
To address modal confusion, some methods like Momentor~\cite{qian2024momentor} and VTG-LLM~\cite{guo2024vtg} introduce extra tokens specifically for time expression, as shown in Fig.~\ref{fig:intro_sub1}b. This approach, however, heavily depends on the classification of these new tokens and the balanced distribution of time data during training. Some tokens may receive insufficient training due to long-tailed distributions in the dataset.
Both approaches share the drawback that using tokens for time expression is inherently discrete, resulting in insufficient precision for time representation (e.g., expressing decimals) and lacking explicit modeling of relationships between adjacent numerical values.
The third approach, exemplified by InternVideo2.5~\cite{wang2025internvideo2} and shown in Fig.~\ref{sec:intro}c, creates trainable task tokens and a specialized temporal perception head to assist in time point decoding. These modules (\eg CG-DETR~\cite{moon2023correlation}) are typically designed for temporal grounding, containing numerous parameters and requiring the re-input of visual information.

To improve the representation of temporal positions in LLMs, we propose \textbf{DisTime}, a method using an additional learnable token to indicate event time spans. This token creates a continuous embedding space for timestamps, thus avoiding confusion with numerical text values. As shown in Fig.~\ref{fig:intro_sub1}{d}, 
DisTime utilizes the \textbf{Distribution-based Time Decoder} to convert the time token into start and end timestamps. 
Inspired by DFL~\cite{li2020generalized}, which acknowledges the inherent ambiguity in boundary detection, temporal boundaries face similar challenges, making direct regression to absolute time values difficult. Decoding the time token into a distribution provides flexibility when event boundaries are unclear, thereby reducing complexity in model prediction. Our method first converts the time token into a probability distribution, and then integrates this distribution to produce specific values. 
Unlike the approach in Fig.~\ref{fig:intro_sub1}c, our decoder utilizes a minimal number of parameters (almost negligible) and without re-inputting original/encoded images.

Another challenge for time-sensitive Video-LLMs is the \emph{scale of temporal-aware datasets}. Current video training datasets~\cite{bain2021frozen,li2024mvbench,yu2019activitynet} predominantly focus on captioning, summarizing, and reasoning, often neglecting temporal awareness. The labor costs of annotating large volumes of time-sensitive data are significant. Although methods like VTimeLLM~\cite{huang2024vtimellm}, InternVid-MR~\cite{gordeev2024saliency}, and Momentor~\cite{qian2024momentor} propose unsupervised approaches to expand temporal grounding data, they share a common limitation: \emph{temporal granularity constraint}, referring to reliance on shot boundaries and coarse fixed temporal intervals.
VTimeLLM uses detected shot boundaries as time points, making event localization prone to errors within single shots, as simple boundaries do not accurately capture event timing. InternVid-MR divides videos into 2-second segments, determining event boundaries based on visual-text similarity, which often results in low labeling accuracy, particularly for short-duration events. Momentor merges segments by assessing the consistency of character instances between consecutive shots, but it still depends on the temporal boundaries of shots, ignoring static objects like scenery.

To expand temporal-aware video data, we propose an annotation paradigm that overcomes temporal granularity constraints for event boundaries. Our method harnesses the captioning abilities of Video-LLMs for event extraction and employs the fine-grained localization capabilities of dedicated models for event boundary detection. Drawing on expertise from well-annotated data, the pseudo labels break free from shot-boundary constraints, making them more suitable for temporal grounding tasks.
We employ three dedicated models: UniMD~\cite{zeng2024unimd} for traditional in-domain temporal grounding, Mr.Blip~\cite{meinardus2024surprising} for high-precision localization with LLMs, and TFVTG~\cite{zheng2024training} for zero-shot settings without training. A scoring strategy evaluates the grounding results, selecting the highest-scoring model output as the ensemble result for each event. Ultimately, we generated 1.25 million event annotations, over 55 times larger than the ActivityNet-Caption~\cite{krishna2017dense} dataset.

To summarize, this paper makes three contributions: 

\begin{itemize}[leftmargin=0.3cm]
\item We propose DisTime, a method that uses a single token for temporal expression in Video-LLMs. 
This approach transforms the token into a distribution via a distribution-based time decoder, which addresses event boundary uncertainty and enables continuous time modeling.

\item We propose an automated annotation paradigm that combines the captioning capabilities of LLMs with the fine-grained grounding abilities of dedicated temporal grounding models. With this paradigm, we create the \textbf{InternVid-TG} dataset, featuring 179K videos and 1.25M events.

\item Experiments demonstrate that DisTime enhances the fine-grained grounding abilities of Video-LLMs across three time-sensitive tasks. Notably, on Charades-STA~\cite{gao2017tall}, our model surpasses all dedicated models and current Video-LLMs in $\text{R@1}_{\text{iou=0.3}}$ score, even in a zero-shot setting.

\end{itemize}

\section{Related Work}
\label{sec:related_work}

\begin{figure*}[ht]
		\centering
		\includegraphics[width=0.85\textwidth]{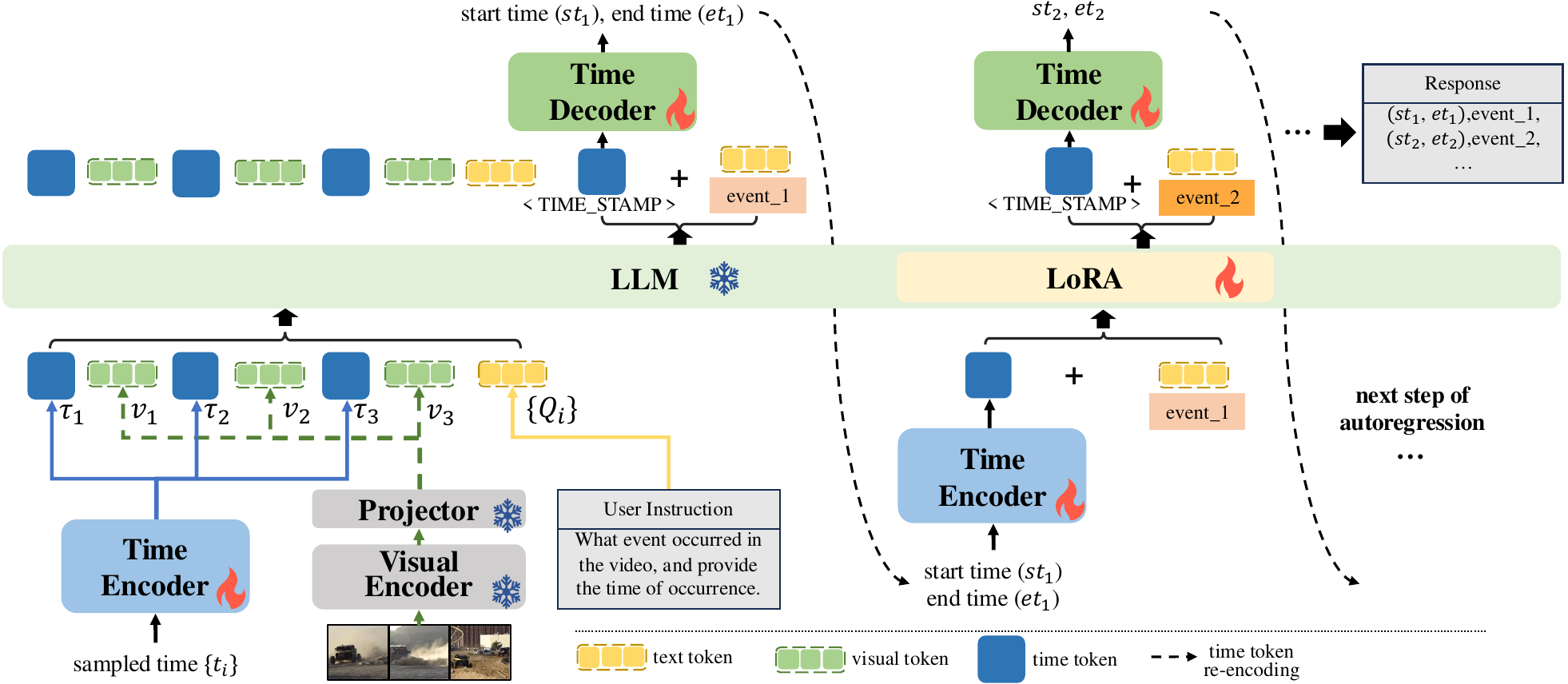}
            \caption{
            \textbf{Overview of the Video-LLM with DisTime.} 
Our DisTime method employs a single token (denoted as $\texttt{<TIME\_STAMP>}$) to represent continuous time, transformed by a distribution-based time decoder and time encoder. 
Initially, sampled timestamps are converted into time tokens and integrated into the LLM's input sequence. During autoregressive generation, when a $\texttt{<TIME\_STAMP>}$ is encountered, its corresponding LLM last-layer embedding is decoded into start and end times. These timestamps are then re-encoded to update the temporal context for subsequent autoregressive steps. In the final output sequence, the decoded times replace the $\texttt{<TIME\_STAMP>}$. This approach enables the Video-LLM to explicitly model time, thereby enhancing its time-sensitive capabilities.
}
\vspace{-5mm}
		\label{demo:network}
\end{figure*}

\noindent \textbf{Time-sensitive video understanding.}
Time-sensitive video understanding tasks require interpreting a video as a sequence of interconnected events and locating them in a temporally grounded manner. Moment Retrieval~\cite{gao2017tall,krishna2017dense,grauman2022ego4d} focuses on identifying the timing of a specific event within a video. Dense Video Captioning~\cite{krishna2017dense,zhou2018towards} involves describing and summarizing all events with their temporal information along the video timeline. Grounded Video Question Answering~\cite{xiao2024can,chen2024cg} demands understanding questions while addressing the embedded temporal information. Thus, these tasks challenge models to have strong temporal sensitivity.

\noindent \textbf{Vision language models.}
With advancements in Large Language Models (LLMs), there is a growing focus on extending their capabilities from natural language processing to the visual domain, enhancing functions such as question answering, description, and summarization based on visual information. 
Recent Vision Language Models~\cite{chen2024far,chiang2023vicuna,li2023blip,liu2023visual,zhu2023minigpt} incorporate learnable modules to integrate language and vision, primarily targeting image-level perception and comprehension. As multimodal research has progressed, Video-LLMs~\cite{zhang2023video,li2024mvbench,li2024llama,li2024llava} have emerged, showing strong performance in tasks like video question answering and captioning. However, these approaches largely concentrate on visual-level task requirements and exhibit limitations in high-level temporal perception.

\noindent \textbf{Temporal-awareness video language models.}
To tackle the challenges of time-sensitive tasks, recent Video-LLMs have focused on constructing time-sensitive data and enhancing temporal awareness in their architectures. GroundingGPT~\cite{li2024groundinggpt} improves performance by generating time-sensitive datasets and integrating multimodal data. VtimeLLM~\cite{huang2024vtimellm} enhances temporal perception through optimized training strategies, enriched datasets, and tailored instruction. TimeChat~\cite{ren2024timechat} and TimeMarker~\cite{chen2024timemarker} incorporate temporal information from the current video frame into the multimodal input, enabling systems to perceive and interpret temporal dynamics effectively. However, these methods model time using textual representation, sharing decision boundaries with regular numerical values, thereby complicating classification and introducing quantization errors.
Momentor~\cite{qian2024momentor} and VTG-LLM~\cite{guo2024vtg} introduce special tokens for timestamp modeling. Momentor uses a temporal perception module to ensure time continuity, while VTG-LLM incorporates absolute-time tokens to manage timestamp information, reducing quantization errors. Despite this, these methods rely heavily on numerous new tokens for representing time absolutely, often overlooking time continuity and boundary fuzziness. We aim to maintain time continuity while addressing boundary fuzziness attentively.
\section{Method}
\label{sec:method}

This section delves into the proposed DisTime, a method that uses a single token to represent continuous time in large language models (LLMs), enabling transformation between time tokens and timestamps via \textbf{Distribution-based Time Encoder-Decoder.}
In Sec.~\ref{method:pipeline}, we present an overview of our approach to time-sensitive video understanding. Sec.~\ref{method:token} details the rationale and strategy for modeling time as a distribution in LLMs. We then introduce the Iterative Time Refinement mechanisms (Sec.~\ref{method:bidirect}) that update the temporal context during the autoregression process. Finally, Sec.~\ref{method:train} outlines the objective function and training methodology for optimizing the time token, time decoder, and time encoder.

\subsection{Overview}
\label{method:pipeline}
The architecture of our time-sensitive video-LLM with DisTime is illustrated in Fig.~\ref{demo:network}, comprising five core components: a visual encoder with a projector, a text encoder, a LLM, a time decoder ($\Phi_{\text{time-dec}}$) and a time encoder ($\Phi_{\text{time-enc}}$).
First, a certain number of video frames are uniformly sampled from the video. Then, the visual encoder encodes the video frames, and the projector maps them into the language space to obtain visual tokens $\{{{{V_{i}}\}}^T_{i=1}}$, where $T$ is the number of sampled time steps in the video. 
Then, we derive a sequence of temporal representations ${\{\tau_i\}^T_{i=1}}$ by processing the timestamps through the time encoder, as:
\vspace{-3mm}
\begin{equation}
\label{eq:time_enc} 
    \tau_i=\Phi_{\text{time-enc}}(t_i), i=1,2, \ldots,T
\end{equation} 
where $t_i\in [0,1]$ is the relative timestamp normalized to the video duration.
Subsequently, the time tokens $\{\tau_i\}_{i=1}^T$ and visual tokens $\{v_i\}_{i=1}^T$ are interleaved and jointly input to the LLM with the user's instruction text tokens $\{Q_i\}_{i=1}^N$, as:
\vspace{-1.5mm}
\begin{equation}
\text{Input} = \text{Concat}({\tau_1, v_1, \dots, \tau_T, v_T}, {Q_1, \dots, Q_N}),
\vspace{-1.5mm}
\end{equation} 
where $\{Q_i\}$ is encoded via text encoder.
We use a dedicated time token, termed \textbf{Distribution-based Time Token}, to represent continuous time. This token is instantiated in the vocabulary as $\texttt{<TIME\_STAMP>}$. 
The autoregressive generation process includes both standard vocabulary tokens and the dedicated time token. When the LLM outputs $\texttt{<TIME\_STAMP>}$, we extract the corresponding hidden state $\mathbf{h}_{time}$ from the final layer and pass it to the time decoder to obtain continuous timestamps, as follows:
\vspace{-2mm}
\begin{equation}
{st, et} = \Phi_{\text{time-dec}}(\mathbf{h}_{time}),
\vspace{-1mm}
\end{equation} 
where $st\in[0,1]$ and $et\in[0,1]$ denote the decoded start time and end time. 
These timestamps are then reprocessed through the time encoder to refresh the temporal context for subsequent autoregressive steps. 
The $\texttt{<TIME\_STAMP>}$ is replaced with the re-encoded time token, establishing explicit temporal referencing that functions as video timeline pointers. 
Finally, the $\texttt{<TIME\_STAMP>}$ is replaced with corresponding decoded timestamps ($st, et$), which are concatenated with generated text tokens (\eg $\texttt{event\_2}$) to form the LLM's complete output. This produces time-sensitive predictions with explicit timestamps.

Our method is compatible with mainstream Video-LLMs (\eg InternVL2.5~\cite{chen2024expanding}, LLaVA-OneVision~\cite{li2024llava}) that retain temporal-aligned input token sequences while being incompatible with the models using global temporal aggregation (\eg LinVT~\cite{gao2024linvt}, BLIP3-Video~\cite{ryoo2024xgen}).

\subsection{Modeling Time as Distribution}

\label{method:token}
\noindent \textbf{DisTime to LLMs.}
Current Video-LLMs excel in global video understanding but often struggle with time-sensitive tasks due to imprecise event timing modeling. Existing solutions either (1) 
express time in text modality (complicating the classification for time and numbers), 
(2) use numerous learnable tokens (suffering from long-tailed distributions and temporal discontinuity), or (3) employ heavy temporal modules (adding extra high computational costs). The representation of time remains a critical optimization target for time-sensitive modeling.

Furthermore, inspired by~\cite{li2020generalized}, we recognize the inherent ambiguity in event boundary detection. For instance, the query ``A person is drinking water from a cup" introduces annotation uncertainty, such as whether to include the cup-picking action in the temporal segment. This ambiguity makes direct regression to absolute timestamps susceptible to precision errors. Holistic analysis of multiple boundary hypotheses is essential to mitigate uncertainty.

To address these limitations, we propose DisTime - a time modeling paradigm that represents time as probability distributions within LLMs. 
Our framework achieves continuous temporal representation through two key innovations:
(1) Single-Token Constraint: This encapsulates continuous temporal information within a dedicated token using a lightweight encoder ($\Phi_{\text{time-enc}}$) and decoder ($\Phi_{\text{time-dec}}$), enabling precise temporal grounding with minimal computational overhead.
(2) Distribution-Aware Decoding: This resolves boundary ambiguity through probabilistic timestamp generation, where final temporal coordinates are derived from weighted combinations of multiple hypotheses.

\begin{figure}[t]
  \centering
  \includegraphics[width=6.8cm]{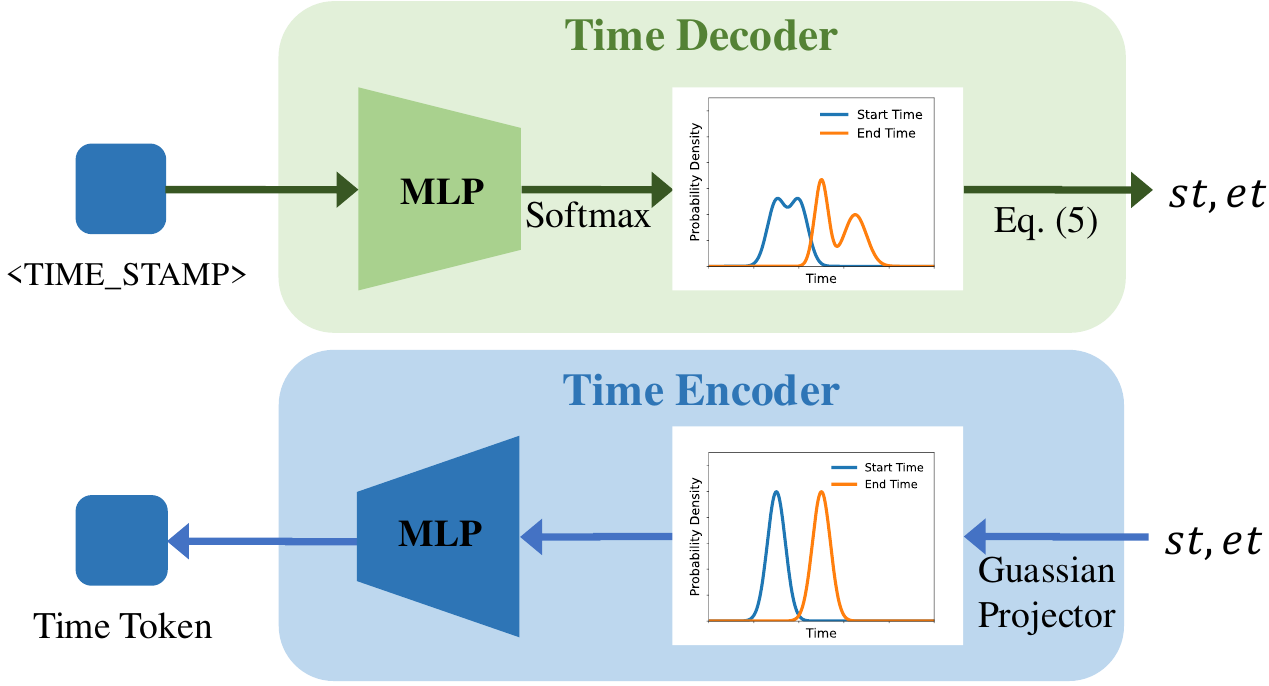}
  \vspace{-2mm}
  \caption{The illustration of time decoder and time encoder.}
  \vspace{-5mm}
  \label{fig:method_enc}
\end{figure}

\noindent \textbf{Distribution-based time token.} 
The dedicated time token in LLMs is termed Distribution-based Time Token (denoted as $\texttt{<TIME\_STAMP>}$), distinctly separating it from numerical text tokens in the vocabulary. When the LLM generates a $\texttt{<TIME\_STAMP>}$, a lightweight decoder with a softmax function transforms it into a latent probability distribution, with each dimension of the latent embedding corresponding to an anchor point on the normalized time axis. These probability distributions enable continuous value representation.

\noindent\textbf{Time decoder.} The time decoder is used to decode the $\texttt{<TIME\_STAMP>}$ into continuous timestamps, as shown in Fig.~\ref{fig:method_enc}.
Specifically, we partition the normalized temporal axis into $reg_{max}+1$ discrete bins,  each linked to an anchor point representing the probability of the timestamp being localized at that position. Formally, the temporal distribution is represented as a vector $\mathbf{e} = [\mathbf{e}_{st}, \mathbf{e}_{et}]$, where $\mathbf{e} \in \mathbb{R}^{2 \times ({reg}_{max}+1)}$, and $\mathbf{e}_{st}$ and $\mathbf{e}_{et}$ denote the start and end time distributions, respectively. A lightweight Multi-Layer Perceptron (MLP) followed by a softmax activation maps the $\texttt{<TIME\_STAMP>}$ to distribution vectors $\mathbf{e}$, as: 
\vspace{-1.5mm}
\begin{equation}
\resizebox{.78\linewidth}{!}{$
\begin{aligned}
\mathbf{e} &  =\Phi_{\text{time-dec}}(\mathbf{h}_{time}) = \text{softmax}(\text{MLP}(\mathbf{h}_{time})). \\
\end{aligned}
$}
\vspace{-1.5mm}
\end{equation}
Continuous timestamps of start time and end time are derived via anchor-weighted summation, as:
\vspace{-1.5mm}
\begin{equation}
st = \textstyle \sum_{i=0}^{reg_{max}} \mathbf{e}_{st}^{(i)} \cdot a_i, \quad  
et = \sum_{i=0}^{reg_{max}} \mathbf{e}_{et}^{(i)} \cdot a_i,
\vspace{-1.5mm}
\end{equation} 
where $\{a_i\}_{i=0}^{reg_{max}}$ are anchors defined as: 
$ a_i = {i}/{reg_{max}}$.

\noindent\textbf{Time encoder.} 
As shown in Fig.~\ref{fig:method_enc}, the time encoder acts as the inverse of the decoder, translating continuous timestamps $(st, et)\in [0,1]$ into a time token ${\tau}$.
Firstly, each timestamp is projected into a Gaussian-regularized distribution to model annotation ambiguity, as: 
$p_{st} \sim \mathcal{N}\big(st, \delta^2\big)$, $p_{et} \sim \mathcal{N}\big(et, \delta^2\big)$,
where $\delta$ controls the distribution spread. Here $\delta=1$ for boundary uncertainty.  
The continuous distributions $p_{st}$ and $p_{et}$ are then discretized into $reg_{max}+1$ uniformly spaced bins spanning $[0,1]$ to obtain the discretized distributions $\mathbf{\hat{e}}_{st}$ and $\mathbf{\hat{e}}_{et}$.
Next, a MLP projects the concatenated distributions $[\mathbf{\hat{e}}_{st}, \mathbf{\hat{e}}_{et}]$ into LLM-compatible time tokens $\tau$, as:
\vspace{-1.5mm}
\begin{equation}
\tau=\text{MLP}([\mathbf{\hat{e}}_{st}, \mathbf{\hat{e}}_{et}]).
\vspace{-1.5mm}
\end{equation}
The encoding process operates through two stages: (1) time token injection during input visual sequence composition, and (2) LLM autoregressive processing. In the former stage, we set $st=et$ to model instantaneous temporal positions. In the latter stage, we utilize timestamps decoded by the time decoder and replace the $\texttt{<TIME\_STAMP>}$ with the encoded token $\tau$ for refined temporal representation.

\subsection{Iterating Time Refinement}
\label{method:bidirect}

\noindent\textbf{Time token re-encoding.} 
The hidden embedding of 
$\texttt{<TIME\_STAMP>}$ inherently capture event boundary ambiguity (\eg multi-modal distributions). 
In the autoregressive generation paradigm of LLMs, retaining raw time tokens across iterations can propagate errors due to representation uncertainty. 
To address this, we use probability-weighted timestamps from the decoder output to refine the LLM's temporal context.
As shown in Fig.~\ref{fig:method_enc}, we re-encode the decoded timestamps into fixed-variance Gaussian distributions via the time encoder. This re-encoded time token serves dual purposes: it transforms ambiguous temporal token interpretations into well-defined Gaussian-distributed representations and ensures distributional alignment across all time tokens. Through this method, arbitrary decoded distributions are normalized into standardized Gaussian forms, enabling consistent temporal representation and enhancing the LLM's temporal comprehension capability.

\begin{figure}[t]
  \centering
  \includegraphics[width=.9\linewidth]{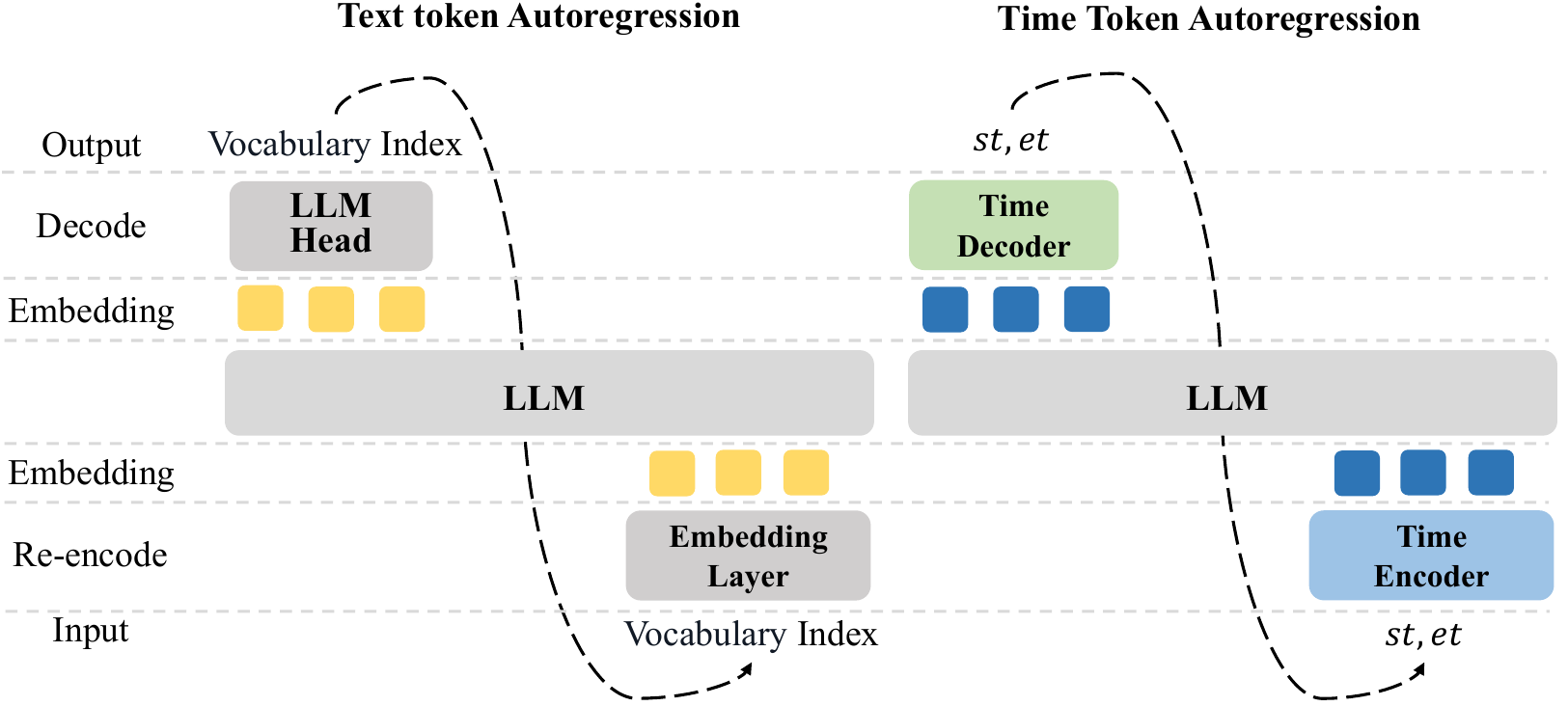}
  \vspace{-2mm}
  \caption{The autoregression of text token and time token.}
  \vspace{-6mm}
  \label{fig:time_ar}
\end{figure}

\noindent\textbf{Time token autoregression.} 
The time token autoregression operates in a similar fashion to text tokens, as depicted in Fig.~\ref{fig:time_ar}. Both token types undergo explicit decoding: text tokens map to vocabulary indices, while time tokens decode to continuous timestamps ($st, et$). The decoded results are then transformed into embeddings for subsequent iterations via domain-specific encoding: an embedding layer for text tokens and a time encoder for time tokens.

\subsection{Training Strategy}
\label{method:train}

\noindent \textbf{Training Object.}
The loss function comprises three components:
distribution focal loss $\mathcal{L}_{dist}$ for temporal distribution learning, 1d-IoU regression loss $\mathcal{L}_{reg}$ to directly optimize temporal grounding, and next token prediction loss $\mathcal{L}_{ntp}$ for autoregressive generation. 
The distribution focal loss~\cite{li2020generalized} is key for training the distribution-based time token and time decoder, as it learns the underlying general distribution without introducing additional priors. 
The unified training objective is formulated as:
$\mathcal{L}=\lambda_1 \cdot \mathcal{L}_{ntp} + \lambda_2 \cdot \mathcal{L}_{reg} + \lambda_3 \cdot \mathcal{L}_{dist}$, where we set $\lambda_1= \lambda_{2}= \lambda_3=1$ for balanced multi-task optimization.

\noindent \textbf{Training Recipe.}
Based on the Video-LLMs that have been trained for general understanding capabilities, our method trains only one stage for enhancing temporal grounding ability. During this stage, the vision backbone and intermediate layers are frozen, and LoRA~\cite{hu2022lora} is employed for efficient fine-tuning of the LLM. Additionally, the parameters of the LLM token embedding, LLM head, and the time encoder and decoder are fully trained.
\section{InternVid-TG}

\begin{figure*}[t]
		\centering
		\includegraphics[width=0.82\textwidth]{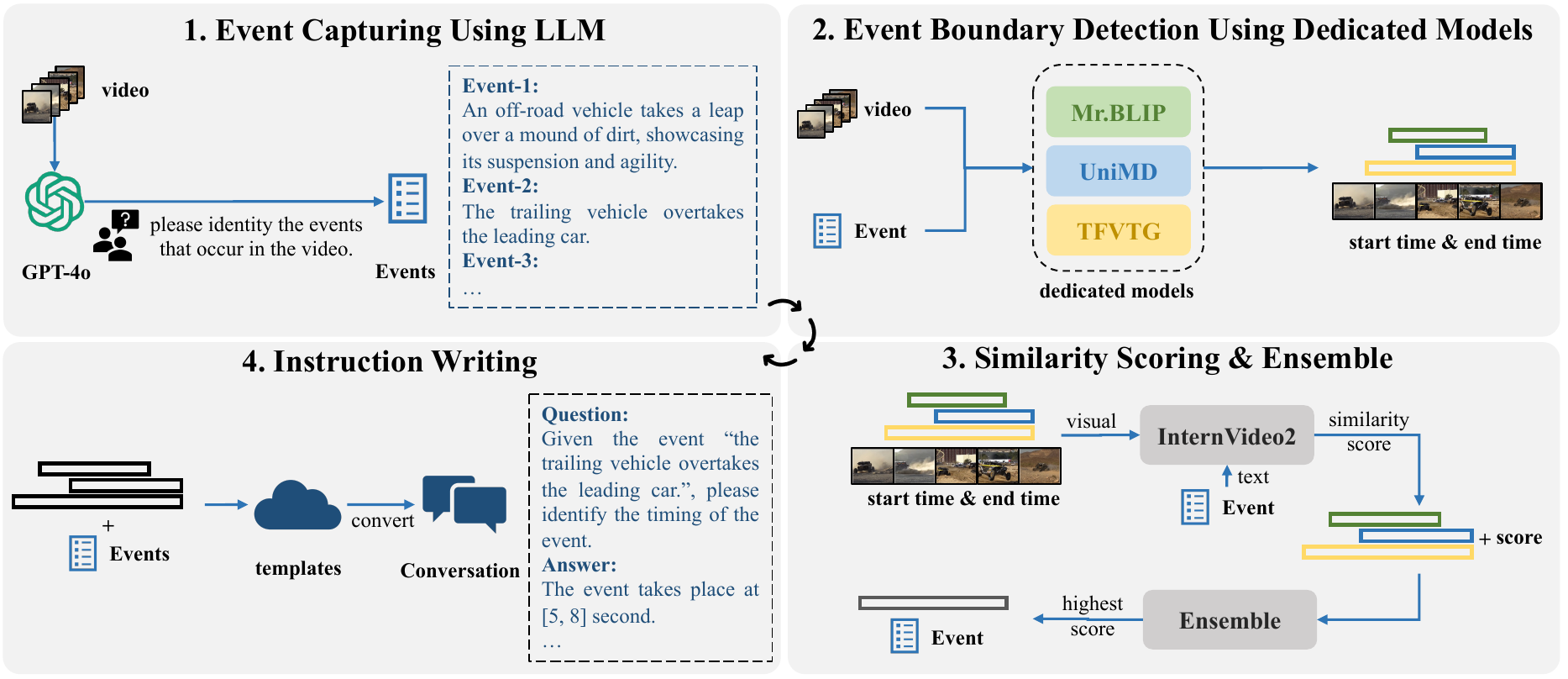}
        \vspace{-2mm}
            \caption{
            The pipeline of our temporal-aware data expansion paradigm, which utilizes multimodal LLM and dedicated temporal models to generate temporal grounding annotations.
            }
            \vspace{-5mm}
		\label{demo:data_process}
\end{figure*}

Existing large-scale video training datasets primarily emphasize captioning, summarization, and reasoning tasks while lacking temporal awareness. Furthermore, though some automated annotation methods~\cite{huang2024vtimellm,gordeev2024saliency,qian2024momentor} attempt to scale supervision, they remain constrained by shot boundaries or coarse fixed temporal intervals. To expand time-aware supervision and overcome these constraints, we propose a temporal-aware data expansion framework that synergistically combines the summarization capabilities of large multimodal models with the fine-grained localization capabilities of temporal dedicated models. 
Fig.~\ref{demo:data_process} illustrates the proposed annotation paradigm, which comprises four steps: event capturing, event boundary detection, scoring and ensemble processes, and instruction writing.
Ultimately, we automate the annotation of InternVid-FLT~\cite{wang2023internvid} video data, transitioning from the original video-text alignment data to temporal grounding data, named \textbf{InternVid-Temporal-Grounding (InternVid-TG)}.

\subsection{Generation pipeline}

\noindent \textbf{Event capturing.}
Video-LLMs are trained on a large amount of video caption data, which endows them with powerful descriptive capabilities. We leverage this to extract event descriptions present in videos.
First, we obtain the untrimmed sources from InternVid-FLT. 
We maintain the original resolution of the video and decode it at 1 fps to form an image sequence, which serves as the visual input for GPT-4o~\cite{achiam2023gpt}. Next, we use the query ``please identify the events that occur in the video'' to prompt the model. In response, GPT-4o generates dense video captions in the format of ``(start time, end time, event description)". Here, we disregard the segment information in the response and use the event description as the extracted event. Finally, we obtain $\approx7.0$ events per video.

\noindent \textbf{Event boundary detection using dedicated models.}
In the step of automatically annotating time points, to eliminate the dependency on shot segmentation, we directly use three models specifically designed for visual temporal grounding to perform the annotation. These models include UniMD~\cite{zeng2024unimd}, Mr.Blip~\cite{meinardus2024surprising}, and TFVTG~\cite{zheng2024training}.

UniMD is a traditional method specialized for temporal grounding. We retrain it using the visual features from InternVideo2~\cite{wang2024internvideo2} as input because the training data for the visual encoder of InternVideo2 includes videos from InternVid-FLT. This adaptation makes UniMD more suited to the domain of InternVid-FLT. 
Mr.Blip is a dedicated Video-LLM trained for temporal grounding, and it is one of the state-of-the-art (SOTA) models in the Charades-STA~\cite{gao2017tall}, ANet-Caption~\cite{krishna2017dense}, and QVHighlight~\cite{lei2021detecting} benchmarks. In QVHighlight, it achieves a top-1 recall at 50\% IoU 
($\text{R@1}_{\text{iou=0.5}}$) 
of 74.77\%, demonstrating a high degree of overlap in event localization.
TFVTG is a train-free video grounding model adapted for zero-shot scenarios. 
In zero-shot situations, it achieves a top-1 recall at 30\% IoU 
($\text{R@1}_{\text{iou=0.3}}$)
of 67.04\% on the Charades-STA benchmark, showing competitiveness with the fine-tuned SOTA LD-DETR~\cite{zhao2025ld} which achieves 73.92\% 
$\text{R@1}_{\text{iou=0.3}}$.

By inputting the extracted event descriptions and the videos, we independently input them into these three dedicated models to obtain the time segment for each event.

\noindent \textbf{Similarity scoring and ensemble.}
To evaluate the segment output from step two, we introduce a similarity scoring strategy. This score subsequently serves as the basis for ensembling the segments from these three models.
We utilize InternVideo2~\cite{wang2025internvideo2} as our scoring model. Initially, each video segment is trimmed, and the video encoder of InternVideo2 is employed to extract visual features. Concurrently, the textual feature of each event caption is derived using InternVideo2's text encoder. The cosine similarity between these features is computed to assess text-video relevance. For each event, the temporal result with the highest score is selected as the ensemble output. 
In the supplementary materials, we present evidence supporting the effectiveness of our scoring strategy and ensemble approach. Additionally, we use the human-annotated InternVid-TG-1k, comprising 1,000 queries, to assess the quality of the pseudo labels.

\begin{table}[t]
\centering
\resizebox{.9\columnwidth}{!}{%
\begin{tabular}{c|ccccc}
\textbf{Dataset}      & annotator & $\text{\#}$videos & Avg Dura. (s) & $\text{\#}$events & $\text{\#}$events/vid \\ \hline
Ego4d~\cite{grauman2022ego4d}        & Human     & 2k      & 472           & ~18.4k  & 9.2         \\
Charades-STA~\cite{gao2017tall} & Human     & 10k     & 30            & ~68k    & 6.8         \\
ANet-caption~\cite{krishna2017dense} & Human     & 15k     & 118           & ~22.5k  & 1.5         \\ \hline
VTimeLLM~\cite{huang2024vtimellm}     & LLM       & 146k    & 52            & ~353k   & 2.4         \\
InternVid-MR~\cite{gordeev2024saliency} & LLM       & -       & -             & ~150k   & -           \\
Momentor~\cite{qian2024momentor}     & LLM       & 65k     & 403           & ~1460k  & 22.5        \\ \hline
\begin{tabular}[c]{@{}c@{}}InternVid-TG\\ (ours)\end{tabular} & \begin{tabular}[c]{@{}c@{}}LLM+\\ dedicated\end{tabular} & 179k & 55 & ~1250k & 7.0 \\ \hline
\end{tabular}%
}
\vspace{-3mm}
\caption{
Comparison among InternVid-TG, existing human-annotated datasets, and unsupervised datasets.
}
\label{tab:data_dist}
\vspace{-6mm}
\end{table}

\noindent \textbf{Instruction writing.}
We design 5 dialogue templates to convert temporal grounding annotations into single-turn conversations, which require identifying timestamps for the given event descriptions in the video. 

\subsection{Impact}

\noindent \textbf{Scale.} Tab.~\ref{tab:data_dist} presents our dataset's data distribution. InternVid-TG annotates 1.25M events from 179k videos, originating from the same source as VTimeLLM and InternVid-MR. However, we extract more events, resulting in the largest video count among these datasets. Overall, our method provides a scalable data expansion approach. Since InternVid-FLT includes 4 million videos, we can expand further to incorporate more videos if needed.

\section{Experiment}
\label{sec:exp}

\begin{table}[!t]
\centering
\resizebox{.75\columnwidth}{!}{%
\begin{tabular}{c|c}
\hline
\textbf{Dataset} & \textbf{Task} \\ \hline
VideoChat2~\cite{li2024mvbench} (232k) & General Video Understanding \\
ANet-Caption~\cite{krishna2017dense} (37k) & Moment Retrieval \\
ANet-Caption~\cite{krishna2017dense} (9.8k) & Dense Video Captioning \\
YouCook2~\cite{zhou2018towards} (2k) & Dense Video Captioning \\
ET-Instcuct~\cite{liu2024bench} (126k) & Multi-Event \\ 
\hline
InternVid-TG (1250k)   & Moment Retrieval \\
\hline
\end{tabular}
}
\vspace{-2mm}
\caption{
Details of the training data covering both general video understanding and time-sensitive comprehension.
}
\vspace{-5mm}
\label{tab:training_data_dist}
\end{table}

In this section, we begin by detailing the experimental setup (Sec.~\ref{subsec:imp}) and benchmarks (Sec.~\ref{subsec:benchmark}). Following this, in Sec.~\ref{subsec:ablation}, we conduct ablation studies to examine the time distribution representation, time token re-encoding, and the effectiveness of our proposed dataset. Finally, in Sec.~\ref{subsec:main}, we provide further comparisons with current Video-LLMs for time-sensitive and general QA tasks. Additional ablation studies, qualitative analyses, and further benchmarks are included in the supplementary material.

\subsection{Implementation Details}
\label{subsec:imp}

The proposed DisTime is integrated into two current Video-LLMs, InternVL2.5~\cite{chen2024expanding} and LLaVA-OneVision~\cite{li2024llava}, to enhance its performance on time-sensitive tasks. 
We use InternVL2.5-1B, LLaVA-OneVision-7B, and InternVL2.5-8B as baselines, with ablation studies conducted on InternVL2.5-1B. For visual input, both Video-LLMs sample frames at equal intervals: InternVL2.5 samples 16 frames, while LLaVA-OneVision samples at most 32 frames. The MLP for the time encoder and decoder consists of three fully connected layers with ReLU functions. The time decoder and encoder are parameter-efficient, collectively comprising just 0.36\% of InternVL2.5-1B, 0.34\% of LLaVA-OneVision-7B and 0.84\% of InternVL2.5-8B. The specific training data are detailed in Tab.~\ref{tab:training_data_dist}.

\subsection{Benchmarks}
\label{subsec:benchmark}

The comprehensive evaluation covers three time-sensitive tasks: Moment Retrieval (MR), Dense Video Captioning (DVC), and Grounded Video Question Answering (Grounded-VQA). For MR, we use the Charades-STA~\cite{gao2017tall}, ANet-Caption~\cite{krishna2017dense}, and QVHighlights~\cite{lei2021detecting}. For DVC, we evaluate on ANet-Caption and YouCook2~\cite{zhou2018towards}. For Grounded-VQA, we employ the NExT-GQA~\cite{xiao2024can} dataset. Additionally, we assess our model's performance on general video understanding tasks using MVBench~\cite{li2024mvbench}, Video-MME~\cite{fu2024video}, and LongVideoBench~\cite{wu2025longvideobench}.

\subsection{Ablation Study}
\label{subsec:ablation}

\noindent \textbf{Roles of distribution representation.}
The comparison between directly predicting timestamps (denoted as ``Direct'') and obtaining timestamps through distribution prediction (denoted as ``Dist.'') is presented in Tab.~\ref{tab:reg_type}. Utilizing distribution representation for time enhances the model's performance across all metrics in the MR task's Charades-STA dataset and the DVC task's YouCook2 dataset. Notably, YouCook2 exhibits a significant improvement, with the F1 score increasing from 2.2\% to 16.3\%.

\noindent \textbf{Roles of time token re-encoding.} 
As shown in Tab.~\ref{tab:reg_type}, time token re-encoding (``Re-Enc'') enhances distribution representation effectiveness, particularly in the high-precision metrics required for the MR task on Charades-STA, where 
$\text{R@1}_{\text{iou=0.7}}$ 
improves by 3\%. This underscores the importance of time token re-encoding for high-precision temporal expressions. Furthermore, in the YouCook2 dataset for the DVC task, ``Re-Enc" leads to an increase of 11.6\% in the CIDEr score and a boost of 4.2\% in the F1 score.

\noindent \textbf{The effectiveness of InternVid-TG.}  
We assess our proposed dataset's effectiveness on two MR benchmarks: Charades-STA and QVHighlights. For a fair comparison with VTimeLLM, we used a subset ($\text{InternVid-TG}^\dagger$) that shares overlapping videos with VTimeLLM, totaling  $\approx$90k videos.
As Tab.~\ref{tab:data_val} shown, compared to VTimeLLM and Momentor, the model with $\text{InternVid-TG}^\dagger$ shows improvements on both benchmarks, even though Momentor contains more events than $\text{InternVid-TG}^\dagger$.
Notably, annotation noise can decrease metrics, as seen with Momentor, where the model’s performance on Charades dropped. Finally, utilizing our full InternVid-TG dataset, expanding from 90k to 179k videos, further boosts performance.

\begin{table}[t]
\vspace{-1mm}
\centering
\resizebox{.95\columnwidth}{!}{
\begin{tabular}{ccc|cccc|ccc}
\hline
\multicolumn{3}{c|}{} & \multicolumn{4}{c|}{\textbf{Charades-STA (MR)}} & \multicolumn{3}{c}{\textbf{YouCook2 (DVC)}} \\
\multicolumn{1}{c}{\multirow{-2}{*}{\textbf{Direct}}} &
\multicolumn{1}{c}{\multirow{-2}{*}{\textbf{Dist.}}} & \multicolumn{1}{c|}{\multirow{-2}{*}{\textbf{Re-Enc.}}} & \begin{tabular}[c]{@{}c@{}}R@1\\ (IoU=0.3) \end{tabular} & \begin{tabular}[c]{@{}c@{}}R@1\\ (IoU=0.5)\end{tabular} & \begin{tabular}[c]{@{}c@{}}R@1\\ (IoU=0.7)\end{tabular} & \begin{tabular}[c]{@{}c@{}}mIoU\end{tabular}& \begin{tabular}[c]{@{}c@{}}SODA\_c\end{tabular} & \begin{tabular}[c]{@{}c@{}}CIDEr\end{tabular} & \begin{tabular}[c]{@{}c@{}}F1 Score\end{tabular} \\ \hline
\multicolumn{1}{c}{ \checkmark } &  &  & 76.6 & 51.9 & 24.9 & 49.5  & 0.6 & 0.9 & 2.2 \\

\multicolumn{1}{c}{} & \multicolumn{1}{c}{ \checkmark } & & 77.0 & 53.5 & 26.7 & 50.1 & 0.9 & 4.0 & 16.3 \\

 & \multicolumn{1}{c}{ \checkmark } & \multicolumn{1}{c|}{ \checkmark } & \textbf{78.1} & \textbf{56.3} & \textbf{29.7} & \textbf{51.6} & \textbf{4.2} & \textbf{15.6} & \textbf{20.5}\\
\hline
\end{tabular}
}
\vspace{-2mm}
\caption{
Ablation study of different components in the time tokenizer. ``Direct'' indicates that the decoder directly predicts timestamps, while ``Dist.'' means the decoder first predicts a distribution and then aggregates it into timestamps. ``Re-Enc.'' denotes the use of time token re-encoding for the next step of autoregression.
}
\vspace{-2mm}
\label{tab:reg_type}
\end{table}
\begin{table}[tp]
\centering
\resizebox{.97\columnwidth}{!}{
\begin{tabular}{c|cccc|cccc}
\hline
\multicolumn{1}{c|}{} & \multicolumn{4}{c|}{\textbf{Charades-STA (MR)}} & \multicolumn{4}{c}{\textbf{QVHighlights (MR)}} \\
\multicolumn{1}{c|}{\multirow{-2}{*}{\textbf{Training Data}}} & \begin{tabular}[c]{@{}c@{}}R@1\\ (IoU=0.3) \end{tabular} & \begin{tabular}[c]{@{}c@{}}R@1\\ (IoU=0.5)\end{tabular} & \begin{tabular}[c]{@{}c@{}}R@1\\ (IoU=0.7)\end{tabular} & \begin{tabular}[c]{@{}c@{}}mIoU\end{tabular} &\begin{tabular}[c]{@{}c@{}}R@1\\ (IoU=0.3)\end{tabular} & \begin{tabular}[c]{@{}c@{}}R@1\\ (IoU=0.5)\end{tabular} & mAP@.5 & mAP@.75\\ \hline
\multicolumn{1}{c|}{Baseline} & 77.4 & 51.4 &22.9 &48.7 & 38.7 & 17.7 & 34.8 & 12.5  \\
\multicolumn{1}{c|}{w/ VTimeLLM~\cite{huang2024vtimellm}} & 76.2 & 51.9 & 25.0 & 49.3 & 51.0 & 25.8 & 46.0 & 18.8 \\
\multicolumn{1}{c|}{w/ Momentor~\cite{qian2024momentor}} & 76.6 & 50.6 & 21.6 & 47.8 & 39.7 & 17.8 & 35.1 & 12.5\\ \hline
\multicolumn{1}{c|}{w/ {InternVid-TG}$^\dagger$} & 77.8 & 55.5 & 28.0 & 51.0 & 52.2 & 26.4 & 46.2 & 18.9\\
\multicolumn{1}{c|}{w/ InternVid-TG} & \textbf{78.1} & \textbf{56.3} & \textbf{29.7} & \textbf{51.6} & \textbf{54.1} & \textbf{27.8} & \textbf{47.9} & \textbf{19.2} \\
\hline
\end{tabular}
}
\vspace{-2mm}
\caption{
The effectiveness of the LLM-generating time-sensitive datasets. Baseline refers to the basic training data, which includes general video understanding and time-sensitive datasets excluding the datasets used for comparison. $\text{InternVid-TG}^\dagger$ denotes a subset that includes videos overlapping with those in VTimeLLM.
}
\vspace{-3mm}
\label{tab:data_val}
\end{table}

\subsection{Main Results}
\label{subsec:main}

\begin{table}[t]
\centering
\resizebox{.95\columnwidth}{!}{
\begin{tabular}{c|c|cccc|cccc} \hline
 &  & \multicolumn{4}{c|}{Charades-STA} & \multicolumn{4}{c}{ANet-Caption}  \\

\multicolumn{1}{c|}{\multirow{-2}{*}{Model}} & \multicolumn{1}{c|}{\multirow{-2}{*}{Size}} & \begin{tabular}[c]{@{}c@{}}R@1\\ (IoU=0.3)\end{tabular} & \begin{tabular}[c]{@{}c@{}}R@1\\ (IoU=0.5)\end{tabular} & \begin{tabular}[c]{@{}c@{}}R@1\\ (IoU=0.7)\end{tabular} & \multicolumn{1}{l|}{mIoU} & \begin{tabular}[c]{@{}c@{}}R@1\\ (IoU=0.3)\end{tabular} & \begin{tabular}[c]{@{}c@{}}R@1\\ (IoU=0.5)\end{tabular} & \begin{tabular}[c]{@{}c@{}}R@1\\ (IoU=0.7)\end{tabular} & \multicolumn{1}{l}{mIoU} \\ \hline

\multicolumn{1}{c}{Dedicated} & \multicolumn{5}{c}{} \\ \hline

UniVTG~\cite{lin2023univtg} & - & 72.6 & 60.2 & 38.6 &52.2 & - & - & - & - \\
UniMD~\cite{zeng2024unimd} & - & - & 63.9 & 42.2 & - & - & 42.7 & 25.1 & - \\
Mr.BLIP~\cite{meinardus2024surprising} & 3B & - & \underline{69.3} & \underline{49.3} & \underline{58.6} & - & \textbf{53.92} & \textbf{35.6} & - \\
LLaVA-MR~\cite{lu2024llava} & 3B & - & \textbf{70.7} & \textbf{49.6} & \textbf{59.8} & - & - & - & -  \\ \hline

\multicolumn{1}{c}{Video-LLMs} & \multicolumn{5}{c}{} \\ \hline

Momentor~\cite{qian2024momentor} & 7B & 42.6 & 26.6 & 11.6 &28.5 & 42.9 & 23.0 & 12.4 & 29.3 \\

ChatVTG~\cite{qu2024chatvtg} & 7B & 52.7 & 33.0 & 15.9 & 34.9 & 40.7 & 22.5 & 9.4 & 27.2  \\

VTimeLLM~\cite{huang2024vtimellm} & 13B & 55.3 & 34.3 & 14.7 &34.6 & 44.8 & 29.5 & 14.2 & 31.4 \\

TimeChat~\cite{ren2024timechat} & 7B & - & 32.2 & 13.4 & - & - & - & - & -  \\

VTG-LLM~\cite{guo2024vtg} & 7B & - & 33.8 & 15.7 & - & - & - & - & - \\

InternVideo2.5~\cite{wang2025internvideo2} & 7B & - & 43.3 & - & 41.7 & - & - & - & - \\ 

TimeMarker~\cite{chen2024timemarker} & 8B & 73.5 & 51.9 & 26.9 & 48.4  & \underline{67.4} & 50.7 & \underline{33.0} & \underline{49.5} \\

InternVL2.5*~\cite{chen2024expanding} & 1B &   3.1 &   1.5 &   0.7  &   5.1 &   5.3 &2.9 & 1.5 & 5.5 \\

LLaVA-OneVision*~\cite{li2024llava} & 7B & 20.8 & 7.2 & 2.2 &  14.5 & 24.8 & 14.0 & 9.0 &  20.5 \\ 

InternVL2.5*~\cite{chen2024expanding} & 8B &   27.9 &   10.9 &   4.2 &   18.2 &   21.1 &   12.7 &   8.6 &   17.2 \\
\hline

\rowcolor{gray!10} DisTime-InternVL & 1B & \underline{78.1} & 56.3 & 29.7 & 51.6 & 67.1 & 45.4 & 23.1 & 45.2  \\ 

\rowcolor{gray!10} DisTime-LLaVAOV & 7B & 76.3 &52.0 & 19.8 & 47.4 & 57.6 & 34.3 & 13.7 & 37.1 \\ 

\rowcolor{gray!10} DisTime-InternVL & 8B & \textbf{81.0} &60.3 & 30.8 & 53.1 & \textbf{72.9} & \underline{53.2} & 30.3 & \textbf{50.5} \\ \hline

\end{tabular}
}
\vspace{-2mm}
\caption{
Comparison with various models on Moment Retrieval task in Charades-STA and ANet-Caption. ``Dedicated'' refers to the model fine-tuned with benchmark-specific training data. ``*'' indicates results from our supplementary testing of the model. Notably, in Charades-STA, our DisTime is in a zero-shot setting.
}
\vspace{-3mm}
\label{tab:mr_res}
\end{table}

\noindent \textbf{Moment Retrieval.}
As shown in Tab.~\ref{tab:mr_res}, our model is compared to dedicated methods and time-sensitive Video-LLMs for the MR task on the Charades-STA, ANet-Caption benchmarks. Our DisTime-InternVL and DisTime-LLaVAOV models outperform all others in $\text{R@1}_{\text{iou=0.3}}$, even in a zero-shot setting. Among Video-LLMs optimized for MR tasks (\ie excluding InternVL2.5 and LLaVA-OneVision), DisTime demonstrates robust time-sensitivity. In Charades-STA, DisTime-InternVL-1B surpasses others across all metrics, notably outperforming TimeMarker by +4.4\% in $\text{R@1}_{\text{iou=0.5}}$. The DisTime-InternVL-8B model further widens this gap to +8.4\%. In ANet-Caption, our 8B model excels in $\text{R@1}_{\text{iou=0.3}}$ and mIoU, trailing slightly behind Mr.BLIP, which is dedicated to ANet-Caption, but remains competitive in $\text{R@1}_{\text{iou=0.5}}$. 
Compared to the baseline InternVL2.5, DisTime significantly enhances the MR task performance. The 1B model demonstrates an improvement in $\text{R@1}_{\text{iou=0.3}}$ on Charades-STA, increasing from 3.1\% to 78.1\%. This validates our method's superior ability to express temporal information in LLMs and to enhance their skills in temporal localization. Furthermore, incorporating DisTime into LLaVA-OneVision considerably increases its temporal sensitivity, showcasing the generalizability and scalability of our method.

\noindent \textbf{Dense Video Captioning.}
As shown in Tab.~\ref{tab:dvc_res}, in the step-wise dense video captioning task, both DisTime-InternVL and DisTime-LLaVAOV significantly outperform all Video-LLMs on the YouCook2 benchmark. This improvement is attributed to our proposed Iterative Time Refinement mechanism, which clarifies the output time token through re-encoding, thereby more accurately conveying the timing of the previous event in each subsequent autoregressive step.
For the ANet-Caption benchmark, our method surpasses Valley and Momentor but falls short of VTimeLLM. It is important to note that, unlike VTimeLLM, which relies on 100-frame inputs, our method uses fewer input frames, with DisTime-InternVL using 16 frames and DisTime-LLaVAOV using 32 frames. Capturing nuanced event progression in ANet-Caption may require richer visual context from additional frames.

\begin{table}[tp]
\centering
\resizebox{.9\columnwidth}{!}{
\begin{tabular}{c|c|ccc|ccc}
\hline
\multicolumn{1}{c|}{\multirow{2}{*}{Model}} & \multicolumn{1}{c|}{\multirow{2}{*}{Size}} & \multicolumn{3}{c|}{YouCook2} & \multicolumn{3}{c}{ANet-Caption} \\
 &  & SODA\_c & CIDEr & \multicolumn{1}{c|}{F1 Score} & SODA\_c & CIDEr & METEOR \\ \hline

Valley~\cite{luo2023valley} & 7B & 0.1 & 0.0 & 1.5 & 0.3 & 1.8 & 0.8 \\

Momentor~\cite{qian2024momentor} & 7B & - & - & - & 2.3 & 14.9 & 4.7 \\

VTimeLLM~\cite{huang2024vtimellm} & 13B & - & - & - & \textbf{5.9} & \textbf{27.2} & \textbf{6.7} \\

TimeChat~\cite{ren2024timechat} & 7B & 1.2 & 3.4 & 12.6 & - & - & - \\

VTG-LLM~\cite{guo2024vtg} & 7B & 1.5 & 5.0 & 17.5 & - & - & - \\ \hline

\rowcolor{gray!10} DisTime-InternVL & 1B & 4.2 & 15.6 & \underline{20.5} & 4.5 & 17.9 & 4.7 \\

\rowcolor{gray!10} DisTime-LLaVAOV & 7B & \underline{5.4} & \underline{21.8} & 19.5 & \underline{4.9} & \underline{21.3} & 4.9 \\ 
  
\rowcolor{gray!10} DisTime-InternVL & 8B & \textbf{6.9} & \textbf{31.0} & \textbf{26.4} & 4.1 & 17.8 & \underline{5.8} \\ \hline

\end{tabular}
}
\vspace{-2mm}
\caption{Comparison with LLM-based models on the Dense Video Captioning task in ANet-Caption and YouCook2.}
\vspace{-6mm}
\label{tab:dvc_res}
\end{table}






\begin{table}[t]
\vspace{-1mm}
\centering
\resizebox{.95\columnwidth}{!}{
\begin{tabular}{c|c|cccccccc}
\hline
\multicolumn{1}{c|}{\multirow{2}{*}{Model}} & \multicolumn{1}{c|}{\multirow{2}{*}{Size}} & \multicolumn{8}{c}{NExT-GQA} \\
\multicolumn{1}{c|}{} & \multicolumn{1}{c|}{} & \multicolumn{1}{c}{Acc@QA} & \multicolumn{1}{c}{Acc@GQA} & \multicolumn{1}{c}{mIoP} & \multicolumn{1}{c}{IoP@0.3} & \multicolumn{1}{c}{IoP@0.5} & \multicolumn{1}{c}{mIoU} & \multicolumn{1}{c}{IoU@0.3} & \multicolumn{1}{c}{IoU@0.5} \\ \hline
SeViLA~\cite{yu2023self} & 3B & 68.1 &16.6 & 25.7 & 34.7 & 22.9 & 21.7 & 29.2 & 13.8 \\
Temp~\cite{xiao2024can} & 150M & 60.2 &16.0 & 25.7 & 31.4 & 25.5 & 12.1 & 17.5 & 8.9 \\
FrozenBiLM~\cite{yang2022zero} & 1B & 70.8  &17.5 & 24.2 & 28.5 & 23.7 & 9.6 & 13.5 & 6.1 \\ \hline

\rowcolor{gray!10} DisTime-InternVL & 1B & 65.9 & 17.0 & 33.5 & 42.9 & 23.7 & 31.1 & 41.0 & 21.8 \\

\rowcolor{gray!10} DisTime-LLaVAOV & 7B & \closeruline{75.8}  & \closeruline{22.8} & \closeruline{36.3} & \closeruline{48.4} & \closeruline{28.7} & \closeruline{31.9} & \closeruline{45.6} & \closeruline{23.5} \\ 

\rowcolor{gray!10} DisTime-InternVL & 8B & \textbf{80.3} & \textbf{28.1} & \textbf{40.5} & \textbf{55.1} & \textbf{33.8} & \textbf{35.9} & \textbf{52.0} & \textbf{28.8}\\

\hline

\end{tabular}}
\vspace{-2mm}
\caption{
Comparison with models on the Grounded Video Question Answering task in NExT-GQA.
}
\vspace{-2mm}
\label{tab:gvqa_res}
\end{table}
\begin{table}[t]
\vspace{-1mm}
\centering
\resizebox{.8\columnwidth}{!}{
\begin{tabular}{c|c|c|c|c}
\hline
\multicolumn{1}{c|}{} & \multicolumn{1}{c|}{} & \multicolumn{1}{c|}{\textbf{MVBench}} & \multicolumn{1}{c|}{\textbf{Video-MME}} & \multicolumn{1}{c}{\textbf{LongVideoBench}}  \\

\multicolumn{1}{c|}{\multirow{-2}{*}{Model}} & \multicolumn{1}{c|}{\multirow{-2}{*}{Size}} & \begin{tabular}[c]{@{}c@{}}avg \end{tabular} & \begin{tabular}[c]{@{}c@{}}(wo sub))\end{tabular} & \begin{tabular}[c]{@{}c@{}}(val total))\end{tabular}  \\ \hline

 \multicolumn{1}{c|}{TimeChat} & \multicolumn{1}{c|}{7B} & 38.5 & 34.7 & -  \\
 \multicolumn{1}{c|}{TimeMarker} & \multicolumn{1}{c|}{8B} & 67.4 & 57.3 & 56.3 \\ 

InternVideo2.5 & 7B & \textbf{75.7} & \textbf{65.1} & \textbf{60.6}  \\
InternVL2.5 & 1B & 64.3 & 50.3 & 47.9 \\
LLaVA-OneVision & 7B & 56.7 & 58.2 & 56.3\\
InternVL2.5 & 8B & \underline{72.0} & \underline{64.2} & \underline{60.0} \\


\hline

\rowcolor{gray!10} DisTime-InternVL & 1B & 62.6 & 47.4 & 49.0 \\

\rowcolor{gray!10} DisTime-LLaVAOV & \multicolumn{1}{c|}{7B} & 59.0 & 54.4 & 55.4 \\

\rowcolor{gray!10} DisTime-InternVL & \multicolumn{1}{c|}{8B} & \underline{72.0} & 59.6 & \underline{60.0} \\

\hline
\end{tabular}
}
\vspace{-2mm}
\caption{
Comparison with models on the General Video Understanding tasks in short (MVBench and Video-MME) and long (LongVideoBench) video benchmarks.
}
\vspace{-6mm}
\label{tab:vqa}
\end{table}

\noindent \textbf{Grounded Video Question Answering.}
The results for the grounded video question answering task on the NExT-GQA benchmark are shown in Tab.~\ref{tab:gvqa_res}. This task not only evaluates the understanding of question answering but also requires the ability to temporally locate events. Among the compared methods, DisTime-InternVL-1B surpasses them in most metrics (except for Acc@GQA and IoP@0.5). When scaled up to DisTime-InternVL-8B, it achieves the top performance across all metrics.

\noindent \textbf{General Video Understanding.}
The results of the general video understanding task are shown in Tab.~\ref{tab:vqa}. 
Without any targeted optimization of general understanding capabilities,  our methods exhibit diverse performance improvements across various benchmarks. In MVBench, DisTime-LLaVAOV-7B outperforms LLaVA-OneVision by 2.3\%, while DisTime-InternVL-8B maintains consistent performance metrics. In LongVideoBench, DisTime-InternVL-1B shows a improvement of 1.1\%. Despite a decline in performance on VideoMME, both DisTime-InternVL and DisTime-LLaVAOV remains competitive when compared to other models. For example, DisTime-InternVL-8B significantly outpaces TimeMarker with the same parameter size, and DisTime-InternVL-1B surpasses TimeChat, which has a much larger parameter size. 
The integration of DisTime primarily aims to enhance time-sensitive tasks, while still maintaining competitiveness in general understanding tasks.

\section{Conclusion}

This paper proposes DisTime, a lightweight paradigm for enhancing temporal expression in Video-LLMs by using a single token to regress continuous timestamps through distribution prediction. Additionally, we propose an automated annotation paradigm that leverages LLMs and specialized temporal grounding models. Video-LLMs equipped with DisTime achieve improved temporal comprehension through efficient continuous time modeling and scalable data generation, providing practical solutions for applications requiring fine-grained video understanding.

\section*{Acknowledgment}
This work was supported by the grants from the National Natural Science Foundation of China 62372014 and Beijing Natural Science Foundation 4252040.

{
    \small
    \bibliographystyle{ieeenat_fullname}
    \bibliography{main}
}

{
\clearpage
\setcounter{page}{1}

\setcounter{table}{0}
\setcounter{figure}{0}
\setcounter{section}{7}
\renewcommand{\thetable}{\Alph{table}}
\renewcommand{\thefigure}{\Alph{figure}}

\maketitlesupplementary

\section*{Appendix}

In the appendix, we describe (1) additional details about the training strategy (Section~\ref{supp:training_details}), (2) further ablation experiments (Section~\ref{supp:ablation}), (3) additional results for moment retrieval (Section~\ref{supp:qvh_res}), (4) the construction of training data instructions (Section~\ref{supp:instruction}), and (5) qualitative results (Section~\ref{supp:quality_res}). For sections, figures, tables, and equations, we use numbers (\eg Sec. 1) to refer to the main paper and capital letters (\eg Sec. A) to refer to this appendix.

\appendix

\section{Training Details}
\label{supp:training_details}

For DisTime-InternVL, we employed a total batch size of 16 throughout the training process. The AdamW optimizer \cite{loshchilov2017decoupled} was applied with a cosine learning rate decay and an initial warm-up period. During training, we used a single epoch with a learning rate set to \(4 \times 10^{-5}\). The LoRA~\cite{hu2022lora} parameters were configured with \(r = 16\) and \(\alpha = 32\). We complete the model training process on 8 A100 GPUs. The completion time for InternVL2.5-1B is approximately 40 hours, while for InternVL2.5-8B, it amounts to around 61 hours.
For DisTime-LLaVAOV, we employed a total batch size of 512 throughout the training process. During training, we used a single epoch with a learning rate set to \(2 \times 10^{-5}\). The LoRA~\cite{hu2022lora} parameters were configured with \(r = 64\) and \(\alpha = 16\). Other settings are consistent with~\cite{li2024llava}. We complete the model training process on 8 A100 GPUs approximately 50 hours.

\section{More Ablation Studies}
\label{supp:ablation}
\noindent \textbf{Effectiveness of the scoring and ensemble in InternVid-TG.} To validate the effectiveness of the scoring, we explore using the query and video data from the Charades-STA dataset, including 3720 queries. We perform inference on these three models in a zero-shot setting and calculate the score and temporal mean-IoU (mIoU) with the ground truth. Fig.~\ref{fig:effect_score} shows the mIoU results of prediction under various scores, indicating a positive correlation between the score and mIoU. 
Next, we explore how to leverage this score to perform an ensemble on the model results.
First, we use queries from the Charades-STA dataset to predict and score the events, setting a series of offset values to examine the bias in the scoring strategy for each model.  We find that the biases for each model are equal, which allows us to directly consider the temporal result with the highest score as the ensemble output. To directly verify the benefits of this ensemble method, we manually annotate the temporal positions for 1k queries extracted from step 1 (referred to as InternVid-TG-1k). 
Tab.~\ref{tab:ensemble} presents the metrics for independent inference of each model and the ensemble results on the InternVid-TG-1k dataset. Our method achieves a mIoU of 43.97\%, which closely approaches the SOTA performance on public datasets (\eg 46.83\% on ANet-Caption), indicating that the ensemble results enhance each component, resulting in more accurate annotation localization.

\begin{figure}[t]
    \centering
    \includegraphics[width=0.5\columnwidth]{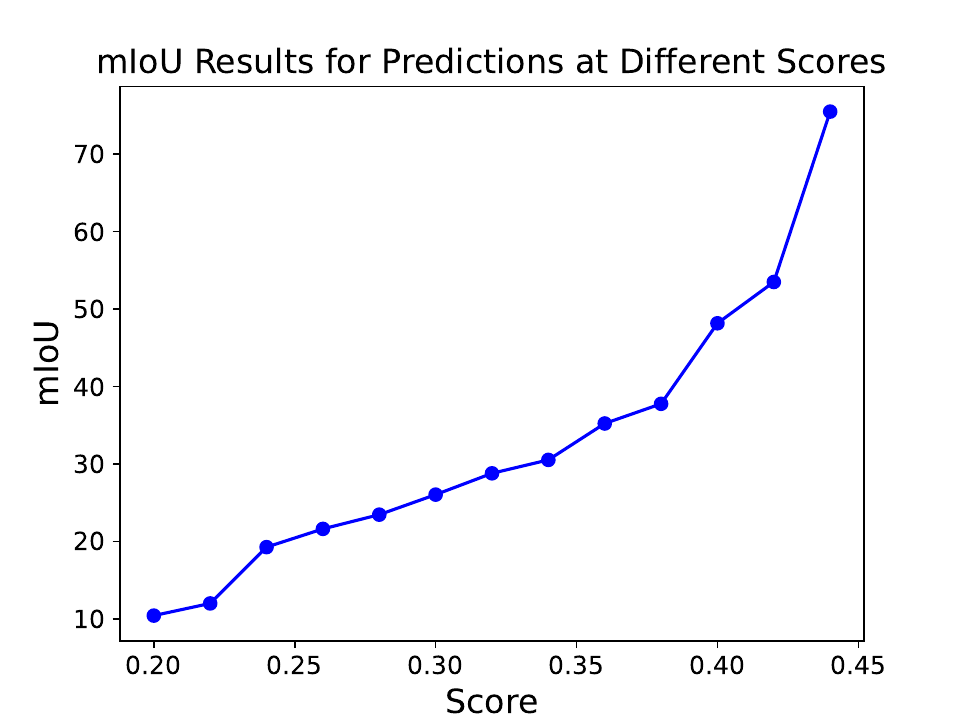}
    \caption{mIoU \vs Confidence Score Curve.}
    \label{fig:effect_score}
\end{figure}

\begin{table}[t]
    \centering
    \resizebox{0.5\columnwidth}{!}{%
        \begin{tabular}{c|ll}
        \hline
        Method   & mIoU  & R1@50 \\ \hline
        UniMD    & 26.63 & 22.22 \\
        Mr.Blip  & 39.81 & 34.34 \\
        TFVTG    & 38.95 & 40.40 \\
        \rowcolor[HTML]{EFEFEF} 
        Ensemble & \textbf{43.97} & \textbf{45.45} \\ \hline
        \end{tabular}
    }
    \caption{Ensemble \vs Individual Model Comparison}
    \label{tab:ensemble}
\end{table}

\noindent \textbf{Number of bins in time decoder.}
The results for different numbers of regression bins in the time decoder are shown in Tab.~\ref{tab:dec_regmax}. As depicted in the table, utilizing 32 bins achieves the best performance in both Charades-STA and YouCook2, except for a slightly lower F1 score compared to using 16 bins (20.5\% \vs 20.9\%). Notably, using a larger number of 64 bins does not result in more accurate time predictions. We believe the large number of bins complicates the model's ability to capture the flow of event timings smoothly.

\begin{table}[ht]
\centering
\resizebox{\columnwidth}{!}{
\begin{tabular}{c|cccc|ccc}
\hline
\multicolumn{1}{c|}{} & \multicolumn{4}{c|}{\textbf{Charades-STA (MR)}} & \multicolumn{3}{c}{\textbf{YouCook2 (DVC)}} \\
\multicolumn{1}{c|}{\multirow{-2}{*}{$\text{\#}reg_{max}$}} & \begin{tabular}[c]{@{}c@{}}R@1\\ (IoU=0.3) \end{tabular} & \begin{tabular}[c]{@{}c@{}}R@1\\ (IoU=0.5)\end{tabular} & \begin{tabular}[c]{@{}c@{}}R@1\\ (IoU=0.7)\end{tabular} & \begin{tabular}[c]{@{}c@{}}mIoU\end{tabular}& \begin{tabular}[c]{@{}c@{}}SODA\_c\end{tabular} & \begin{tabular}[c]{@{}c@{}}CIDEr\end{tabular} & \begin{tabular}[c]{@{}c@{}}F1 Score\end{tabular} \\ \hline
\multicolumn{1}{c|}{16} & 77.6 & 54.4 & 27.6 & 50.2 & 4.0 & 12.1 & \textbf{20.9} \\

 \multicolumn{1}{c|}{32}  &\textbf{78.1} & \textbf{56.3} & \textbf{29.7} & \textbf{51.6} & \textbf{4.2} & \textbf{15.6} & 20.5  \\
 
 \multicolumn{1}{c|}{64}  & 77.0 & 53.7 & 27.4 & 50.3 & 2.9 & 11.2 & 16.4 \\
\hline
\end{tabular}
}
\caption{Ablation study on the number of bins ($reg_{max}$) in time decoder.}
\label{tab:dec_regmax}
\end{table}


\noindent \textbf{Number of layers in time decoder and encoder.}
As shown in Tab.~\ref{tab:mlp_layers}, the number of time decoder and encoder layers significantly impacts the final results. Utilizing three layers yields optimal performance, while further increasing the number of layers does not lead to better results. We believe that the embeddings produced by the LLM inherently contain substantial temporal information, and excessive layers may disrupt this intrinsic data.
\begin{table}[t]
\centering
\resizebox{\columnwidth}{!}{
\begin{tabular}{c|cccc|ccc}
\hline
\multicolumn{1}{c|}{} & \multicolumn{4}{c|}{\textbf{Charades-STA (MR)}} & \multicolumn{3}{c}{\textbf{YouCook2 (DVC)}} \\
\multicolumn{1}{c|}{\multirow{-2}{*}{$\text{\#}layers$}} & \begin{tabular}[c]{@{}c@{}}R@1\\ (IoU=0.3) \end{tabular} & \begin{tabular}[c]{@{}c@{}}R@1\\ (IoU=0.5)\end{tabular} & \begin{tabular}[c]{@{}c@{}}R@1\\ (IoU=0.7)\end{tabular} & \begin{tabular}[c]{@{}c@{}}mIoU\end{tabular}& \begin{tabular}[c]{@{}c@{}}SODA\_c\end{tabular} & \begin{tabular}[c]{@{}c@{}}CIDEr\end{tabular} & \begin{tabular}[c]{@{}c@{}}F1 Score\end{tabular} \\ \hline
\multicolumn{1}{c|}{2} & 77.1 & 52.4 & 26.6 & 50.1 & 3.4 & 10.1 & 17.0 \\

 \multicolumn{1}{c|}{3}  &\textbf{78.1} & \textbf{56.3} & \textbf{29.7} & \textbf{51.6} & \textbf{4.2} & \textbf{15.6} &\textbf{20.5}  \\
 
 \multicolumn{1}{c|}{4} & 77.3 & 54.4 & 27.9 & 50.0 & 3.6 & 15.5 & 18.0 \\
\hline
\end{tabular}
}
\caption{Ablation study on the number of layers ($L$) in time decoder and time encoder.}
\label{tab:mlp_layers}
\end{table}

\section{More Results}
\label{supp:qvh_res}
\noindent \textbf{Moment retrieval task on QVHighlights.}
As shown in Tab.~\ref{tab:qvh_res}, we surpass previous video-LLMs by a large margin ($7.6\% \xrightarrow{}53.8\%$ in mAP@0.5). However, we still lag behind dedicated models. This is due to the presence of multiple segments for a single event in the QVHighlight annotations, which makes it challenging for video-LLM models to recall the targets effectively.

\begin{table}[t]
\centering
\resizebox{.95\columnwidth}{!}{
\begin{tabular}{c|c|cccc}
\hline
 &  & \multicolumn{4}{c}{QVHighlights}   \\
\multicolumn{1}{c|}{\multirow{-2}{*}{Model}} & \multicolumn{1}{c|}{\multirow{-2}{*}{Size}} & \begin{tabular}[c]{@{}c@{}}R@1\\ (IoU=0.3)\end{tabular} & \begin{tabular}[c]{@{}c@{}}R@1\\ (IoU=0.5)\end{tabular} & mAP@.5 & mAP@.75\\ \hline

\multicolumn{1}{c}{Dedicated} & \multicolumn{5}{c}{} \\ \hline

Mr.BLIP~\cite{meinardus2024surprising} & 3B & \underline{74.8} & \underline{60.5} & \underline{68.1} &{\underline{53.4}}  \\
LLaVA-MR~\cite{lu2024llava} & 3B & \textbf{76.6} & \textbf{61.5} & \multicolumn{1}{c}{\textbf{69.4}} & \textbf{54.4} \\ \hline

\multicolumn{1}{c}{Video-LLMs} & \multicolumn{5}{c}{} \\ \hline

Momentor~\cite{qian2024momentor} & 7B & 17.0 & - & 7.6 & -  \\

\rowcolor{gray!10} DisTime-InternVL & 1B & 54.1 & 27.8 & 47.9 & 19.2  \\ 

\rowcolor{gray!10} DisTime-LLaVAOV & 7B & 44.1 & 14.9 & 37.9 & 8.4 \\ 

\rowcolor{gray!10} DisTime-InternVL & 8B & 61.1 &37.5 & 53.8 & 28.1 \\ 

\hline

\end{tabular}
}
\caption{
Comparison with different models on moment retrieval task in QVHighlights. ``Dedicated'' refers to the model fine-tuned with benchmark-specific training data. Notably, our DisTime is in a zero-shot setting.
}
\label{tab:qvh_res}
\end{table}

\section{Instructions for Time-Sensitive Task}
\label{supp:instruction}
In addition to the rich instruction data included in the ET-Instruct dataset~\cite{liu2024bench}, we expanded the instructions for the moment retrieval (used in InternVid-TG) and the dense video captioning. Specifically, we referred to TimeChat~\cite{ren2024timechat} and ET-Instruct, used GPT-4o~\cite{achiam2023gpt} to expand more high-quality instructions, and finally manually selected the generated instructions as the final templates. Tab.~\ref{tab:instruction} shows our instruction examples, answer format, and output examples for the moment retrieval and dense video captioning.

\definecolor{lightcyan}{RGB}{255,250,205}
\definecolor{mediumpurple}{RGB}{230,230,250}

\begin{table*}[htp]
\centering
\resizebox{\textwidth}{!}{
\begin{tabular}{l|l|l}
\hline
\textbf{Task} & \textbf{Type} & \textbf{Example} \\ \hline
\multirow{13}{*}{Moment Retrieval} & Instruction Example & \begin{tabular}[c]{@{}l@{}}Give you a textual query: \colorbox{lightcyan}{\textless{}query\_placeholder\textgreater{}}. When does the described content occur in the video? Please return the timestamp.\\ 
Here is a text query:\colorbox{lightcyan}{\textless{}query\_placeholder\textgreater{}}. At what point in the video does the described event happen? Please provide the timestamp.\\ 
Analyze the event description:\colorbox{lightcyan}{\textless{}query\_placeholder\textgreater{}}. At what moment in the video does the described event take place? Return the timestamp.\\
Consider the query: \colorbox{lightcyan}{\textless{}query\_placeholder\textgreater{}}. When is the described event occurring in the video? Kindly provide the timestamp.\\ 
Examine the following text query: \colorbox{lightcyan}{\textless{}query\_placeholder\textgreater{}}. When is the described event taking place in the video? Please return the timestamp.\end{tabular} \\
\cline{2-3} 

& Answer Format & \begin{tabular}[c]{@{}l@{}}The event occurs at \colorbox{mediumpurple}{\textless{}TIME\_STAMP\textgreater{}}.\\ 
The described event takes place at \colorbox{mediumpurple}{\textless{}TIME\_STAMP\textgreater{}}.\\ 
This situation happens at \colorbox{mediumpurple}{\textless{}TIME\_STAMP\textgreater{}}.\\ 
This event is at \colorbox{mediumpurple}{\textless{}TIME\_STAMP\textgreater{}}.\\ 
It takes place at \colorbox{mediumpurple}{\textless{}TIME\_STAMP\textgreater{}}.\end{tabular} \\ 
\cline{2-3} 

& Output Example & \begin{tabular}[c]{@{}l@{}}The event occurs at 1.2s - 5.8s.\\ The described event takes place at 3.4s - 7.2s.\\ This situation happens at 22.1s - 36.0s.\\ This event is at 12.5s - 30.1s.\\ It takes place at 33.1 - 41.0s.\end{tabular} \\ \hline
\multirow{12}{*}{Dense Video Caption} & Instruction Example & \begin{tabular}[c]{@{}l@{}}Identify and localize a series of steps or actions occurring in the video, providing start and end timestamps and related descriptions.\\ Localize a series of action steps in the given video, output a start and end timestamp for each step, and briefly describe the step.\\ Capture and describe the activity events in the given video, specifying their respective time intervals, and output the time.\\ Pinpoint the time intervals of activity events in the video, and provide detailed descriptions for each event.\\ Detect and report the start and end timestamps of activity events in the video, along with descriptions.\end{tabular} \\ \cline{2-3} 
 & Answer Format & \begin{tabular}[c]{@{}l@{}}\colorbox{mediumpurple}{\textless{}TIME\_STAMP\textgreater{}}, Step1.\colorbox{mediumpurple}{\textless{}TIME\_STAMP\textgreater{}}, Step2. \colorbox{mediumpurple}{\textless{}TIME\_STAMP\textgreater{}}, Step3...\\
 \colorbox{mediumpurple}{\textless{}TIME\_STAMP\textgreater{}}, Step1.\colorbox{mediumpurple}{\textless{}TIME\_STAMP\textgreater{}}, Step2. \colorbox{mediumpurple}{\textless{}TIME\_STAMP\textgreater{}}, Step3...\\\colorbox{mediumpurple}{\textless{}TIME\_STAMP\textgreater{}}, Event1.\colorbox{mediumpurple}{\textless{}TIME\_STAMP\textgreater{}}, Event2. \colorbox{mediumpurple}{\textless{}TIME\_STAMP\textgreater{}}, Event3 ...\\\colorbox{mediumpurple}{\textless{}TIME\_STAMP\textgreater{}}, Event1.\colorbox{mediumpurple}{\textless{}TIME\_STAMP\textgreater{}}, Event2. \colorbox{mediumpurple}{\textless{}TIME\_STAMP\textgreater{}}, Event3 ... \\ 
\colorbox{mediumpurple}{\textless{}TIME\_STAMP\textgreater{}}, Event1.\colorbox{mediumpurple}{\textless{}TIME\_STAMP\textgreater{}}, Event2. \colorbox{mediumpurple}{\textless{}TIME\_STAMP\textgreater{}}, Event3 ...\end{tabular} \\ \cline{2-3} 
 
& Output Example & \begin{tabular}[c]{@{}l@{}}23.7s - 69.8s, spread the meat on the foil. 74.5s - 111.6s, cut the meat into four pieces. 114.7s - 146.0s, ...\\ 

26.7s - 50.8s, add oil to the wok. 49.3s - 82.2s, add garlic and green onions to the wok.81.7 - 107.7s, add ...\\

2.1s - 96.4s, a large orange leaf blower blows leaves in a yard. 96.5s - 196.8s, a man drives the leaf blower.\\ 

2.5s - 18.1s, a man is seen speaking to the camera and leads into him pouring oil into a pot. 18.2s - 58.8s, he ...\\ 

0.3s - 17.2s, a girl is seen climbing across a set of monkey bars while looking back to the camera. 17.5s - 32.8s, ...\end{tabular} \\ 
\hline
\end{tabular}}
\caption{
Example of instructions for moment retrieval and dense video captioning. \textless{}query\_placeholder\textgreater{} is the query placeholder for moment retrieval, and \textless{}TIME\_STAMP\textgreater{}is the original response of LLM, which will be replaced by the time result decoded by the Time Decoder.}
\label{tab:instruction}
\end{table*}

\section{Qualitative Analyses}
\label{supp:quality_res}

\noindent \textbf{Impact of temporal distribution.}
We demonstrate the importance of temporal distribution in temporal grounding tasks, as illustrated in Fig.~\ref{demo:visTD}. When the model receives the event query, it initially perceives an open cupboard door in the frame at 0 seconds, causing a slight response in the start time distribution curve. However, as time progresses, this response diminishes until the person begins to close the door, at which point the start time responds again. As the door-closing action comes to an end, the response for the end time gradually weakens. In the frame at 30 seconds, a hand blocks the cupboard door, resulting in another slight response from the model. Compared to the Dirac distribution, representing time as a broader distribution is more suitable for capturing events with blurred boundaries.

\begin{figure}[!h]
		\centering
		\includegraphics[width=\columnwidth]{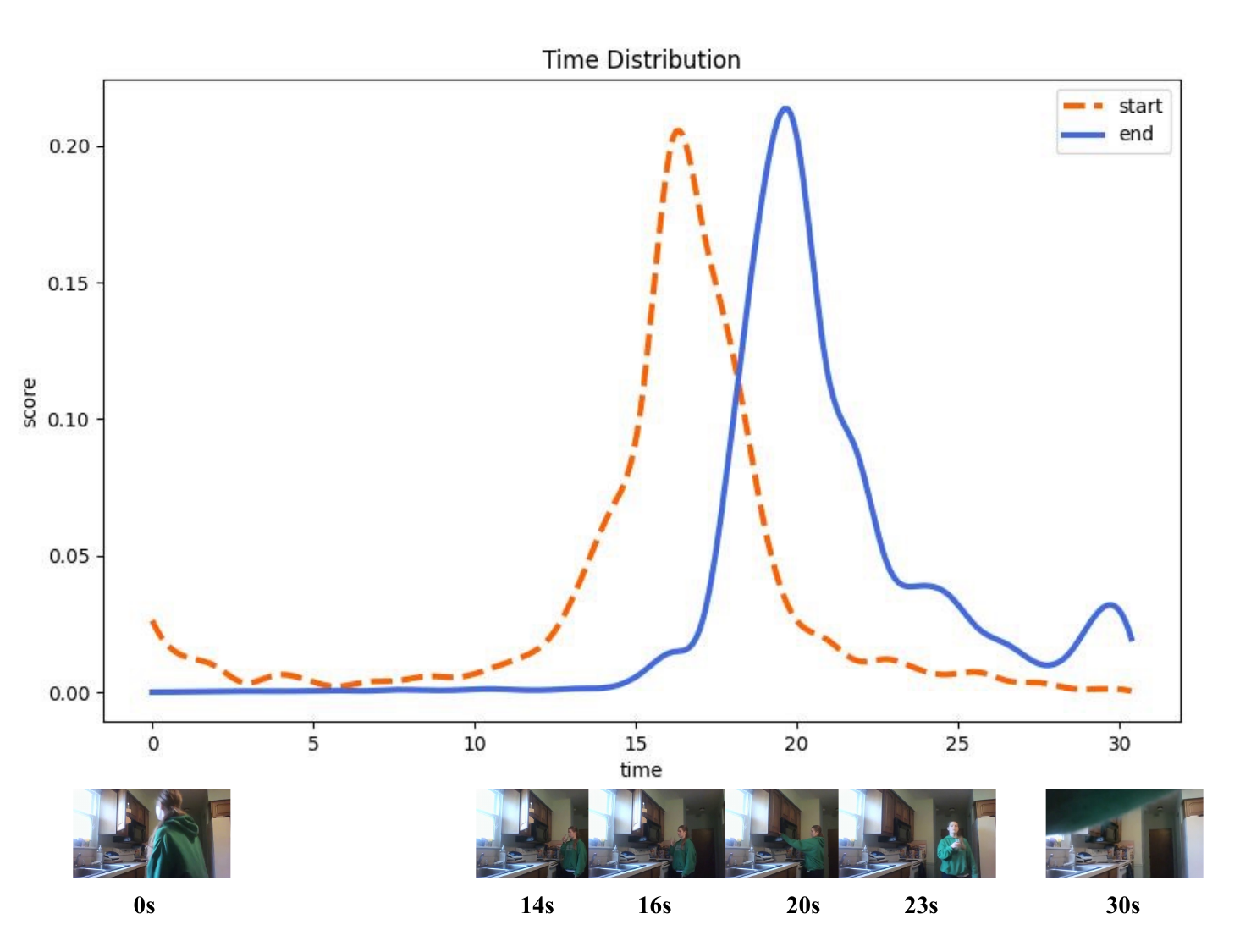}
            \caption{Visualization curves of the start and end time in moment retrieval. Query: person closes a cupboard door. Orange curve represents the start time distribution; Blue curve represents the end time distribution.}
		\label{demo:visTD}
            \vspace{-3.0mm}
    
\end{figure}

\noindent \textbf{Qualitative results.}
This section presents the qualitative results from videos involving multiple tasks, including open-ended question answering and moment retrieval, as illustrated in Fig.~\ref{demo:visQA}. Additionally, we provide a visual comparison between DisTime’s predictions and the ground truth on temporal grounding tasks. Specifically, for moment retrieval, we show the visualization results of DisTime on the Charades-STA and ANet-Caption, as depicted in Fig.~\ref{demo:visMR}. For dense video captioning, we display the visualization results of DisTime on the step description dataset YouCook2 and the event description dataset ANet-Caption, as illustrated in Fig.~\ref{demo:visYoucook2-DVC} and Fig.~\ref{demo:visANet-DVC}, respectively. This comparison demonstrates the advanced capabilities of our method in accurately modeling and predicting temporal events.

\begin{figure*}[t]
		\centering
		\includegraphics[width=\textwidth]{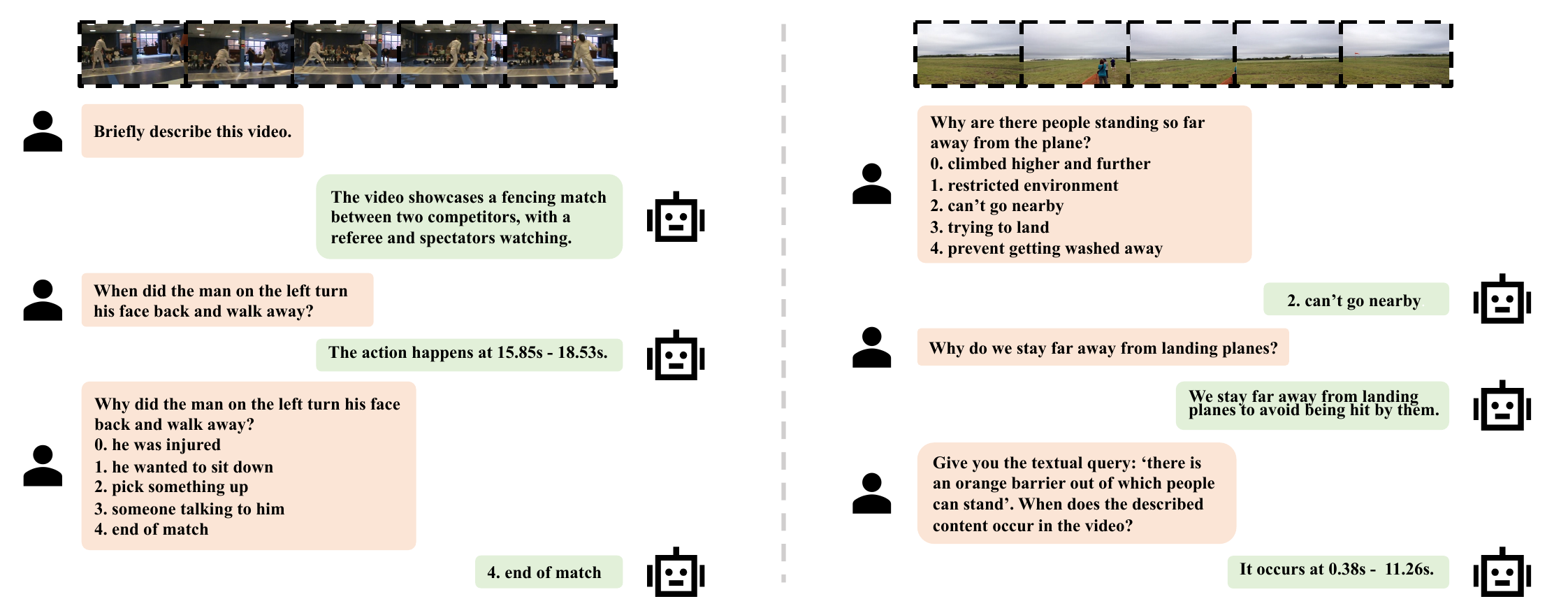}
        \vspace{-5mm}
            \caption{Qualitative results on multiple different tasks, such as temporal grounding task, video question answer, and open-ended question answer.}
		\label{demo:visQA}
            \vspace{-3.0mm}
    
\end{figure*}
\begin{figure*}[t]
    \centering
    \subfloat[Qualitative results on Charades-STA.\label{subfig:a}]{
        \includegraphics[width=.95\textwidth]{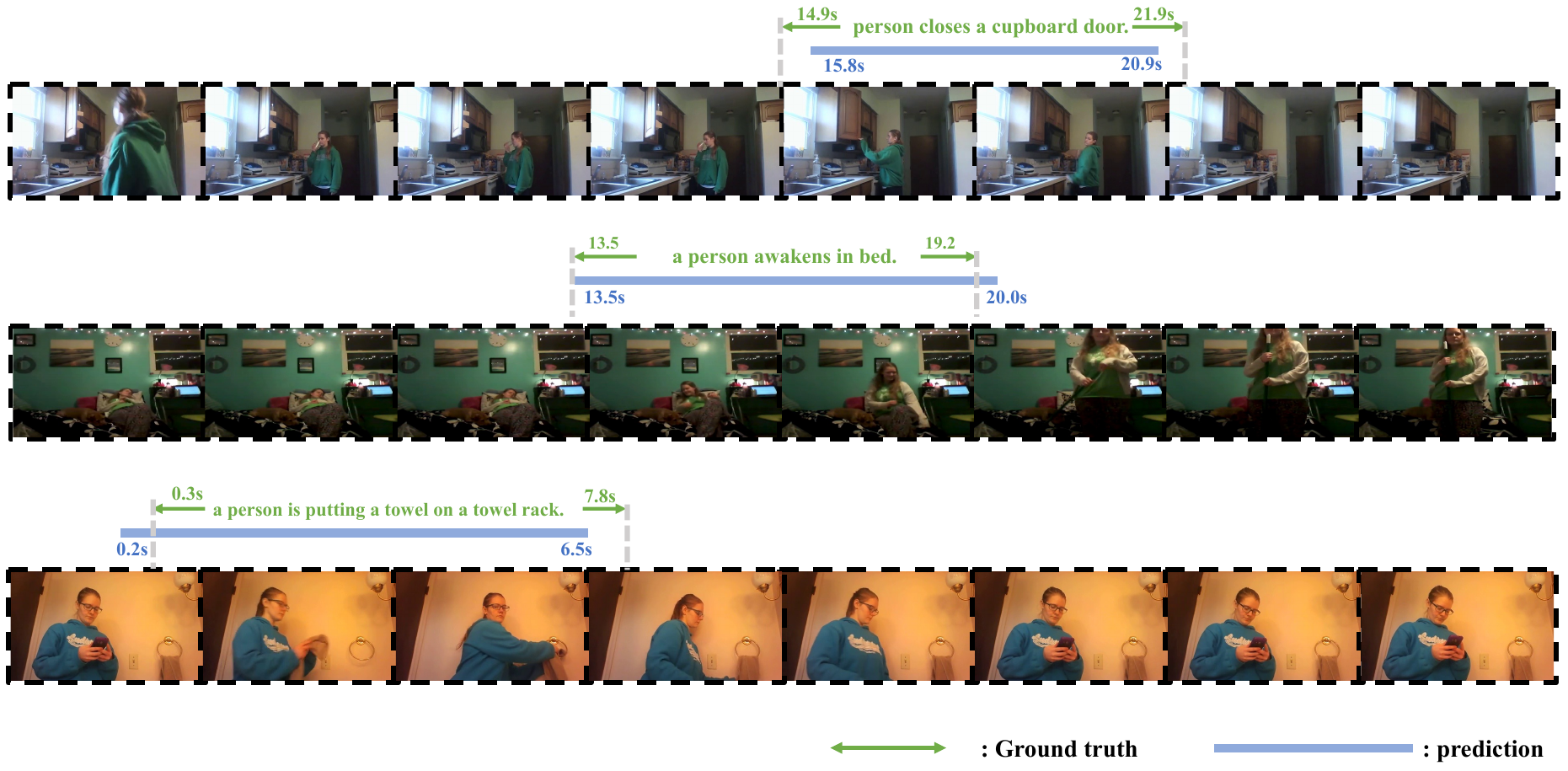}
    }\qquad
    \subfloat[Qualitative results on ANet-Caption.\label{subfig:b}]{
        \includegraphics[width=.95\textwidth]{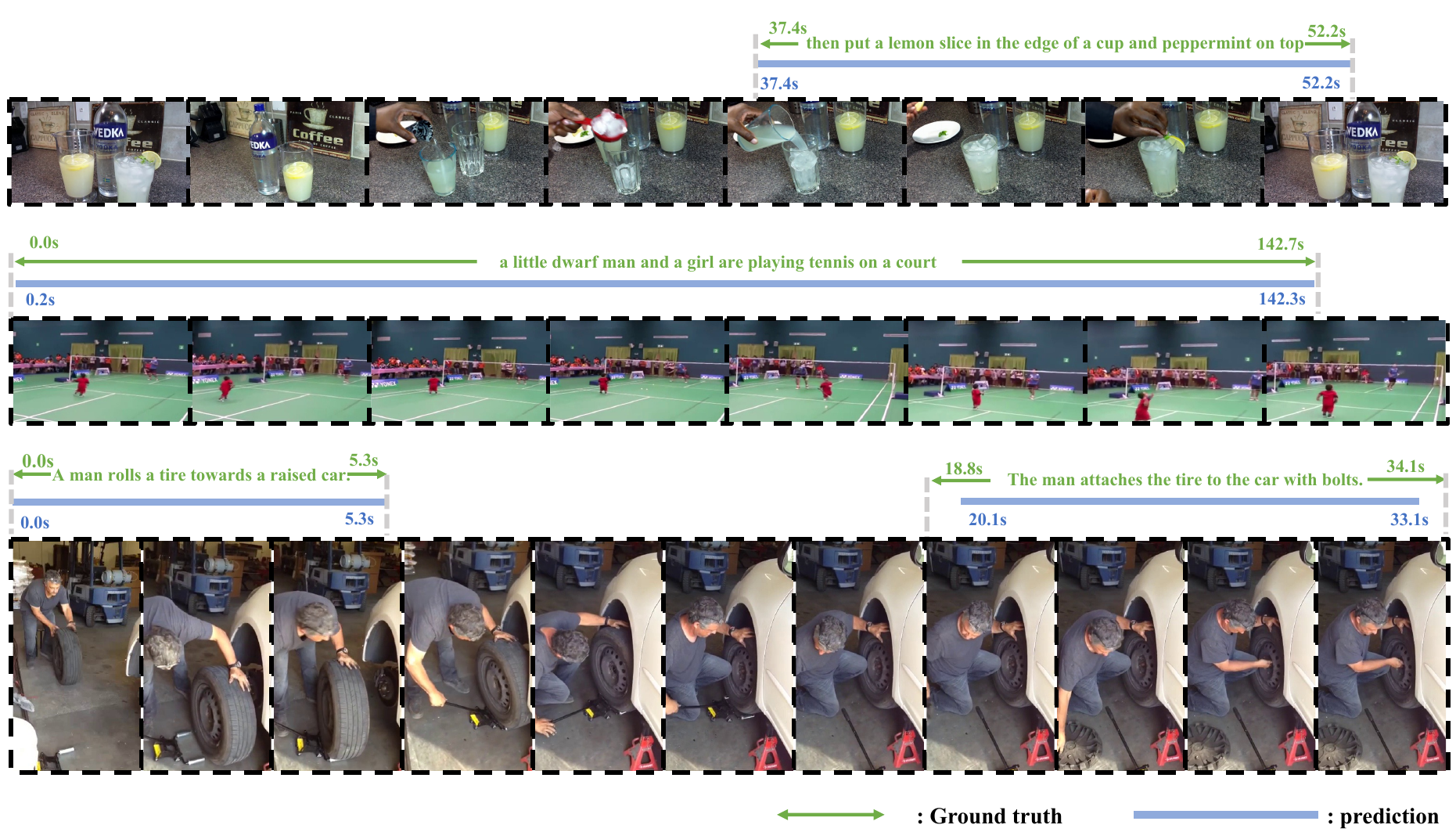}
    }
    \caption{Visual comparison of ground truth and prediction of moment retrieval. The arrow indicates the ground truth, and the rectangular progress bar indicates the prediction of DisTime.}
    \label{demo:visMR}
    \vspace{-3.0mm}
\end{figure*}

\begin{figure*}[t]
		\centering
		\includegraphics[width=\textwidth]{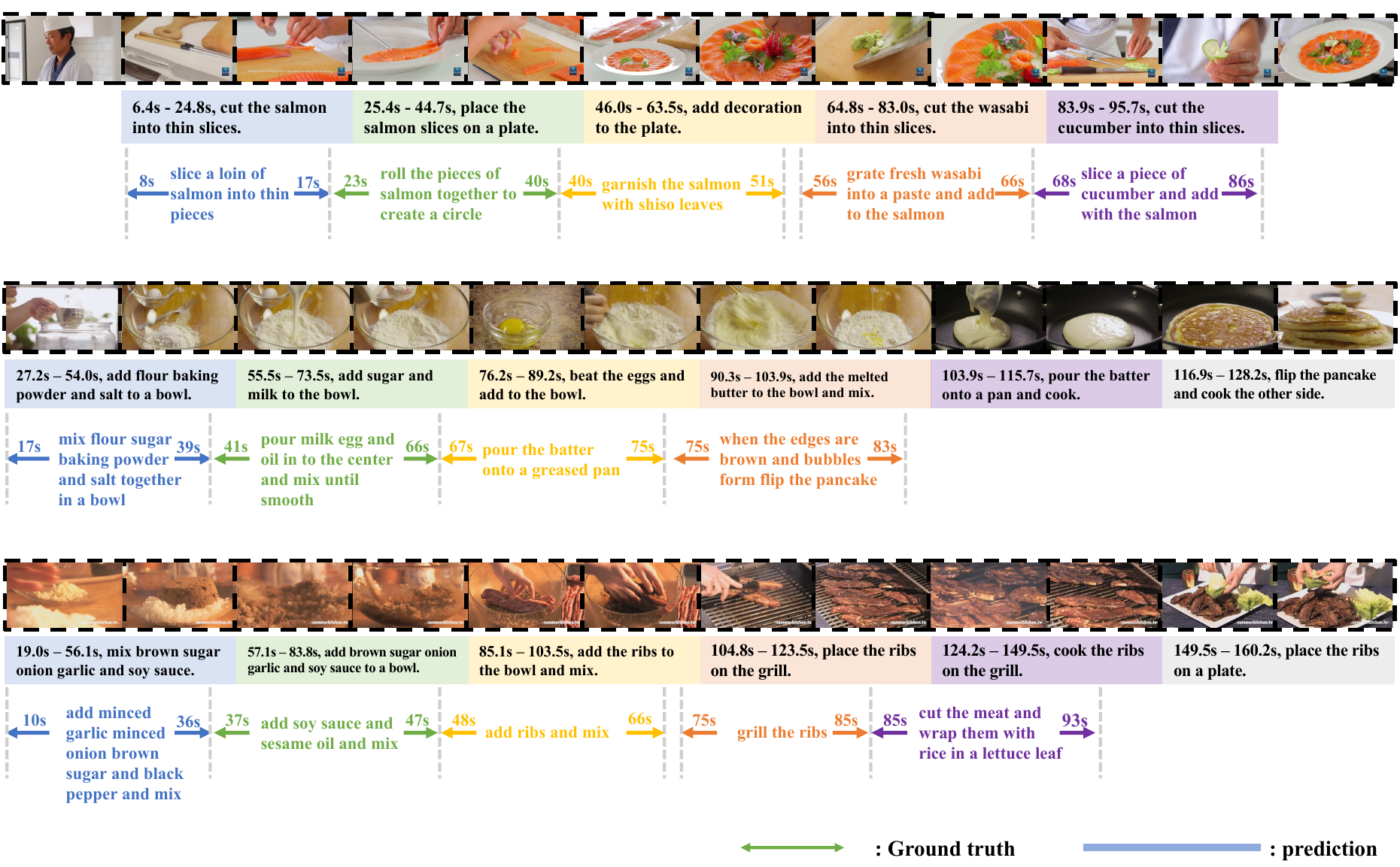}
        \vspace{-5mm}
            \caption{Visualization of ground truth and prediction from YouCook2. The arrow indicates the ground truth, and the rectangular progress bar indicates the prediction of DisTime.}
		\label{demo:visYoucook2-DVC}    
\end{figure*}

\begin{figure*}[t]
		\centering
		\includegraphics[width=\textwidth]{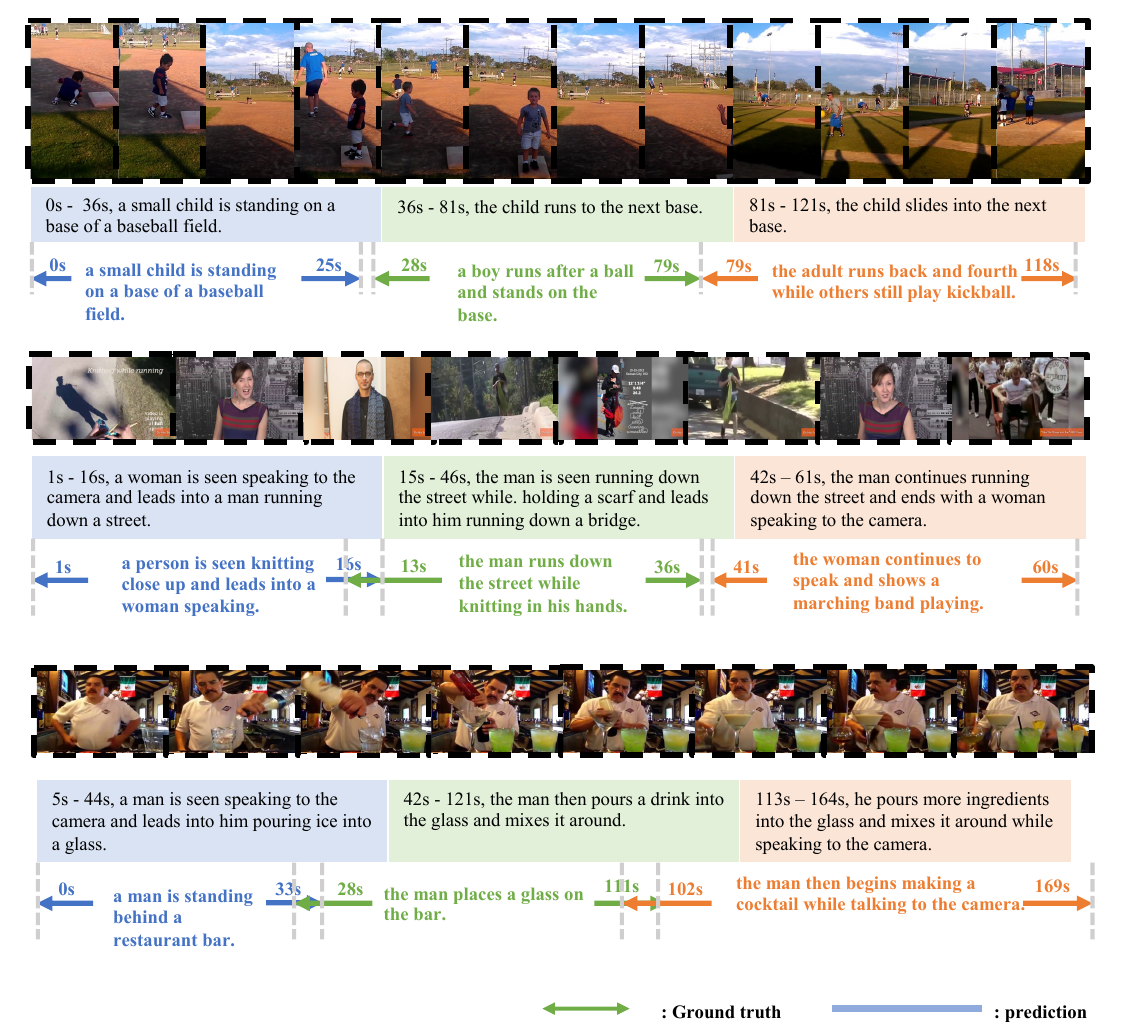}
        \vspace{-5mm}
            \caption{Visualization of ground truth and prediction from ANet-Caption. The arrow indicates the ground truth, and the rectangular progress bar indicates the prediction of DisTime.}
		\label{demo:visANet-DVC}
            \vspace{-3.0mm}
\end{figure*}
}

\end{document}